\documentclass[preprints,article,accept,moreauthors,pdftex]{Definitions/mdpi} 
\usepackage{subfigure}

\firstpage{1} 
\pubvolume{xx}
\issuenum{1}
\articlenumber{5}
\pubyear{2020}
\copyrightyear{2020}
\history{}
 
\Title{{\spaceskip=0.53em\relax A Framework of Combining Short-Term Spatial/}\\ {\spaceskip=0.2em\relax Frequency Feature Extraction and Long-Term IndRNN} for Activity Recognition}

\Author{Beidi Zhao $^{1, \dagger}$, Shuai Li $^{2,\dagger, }$*, Yanbo Gao $^{2}$, Chuankun Li$^3$, Wanqing Li$^4$}

\AuthorNames{Beidi Zhao $^{1,\ddagger}$\orcidA{}, Shuai Li $^{2,\ddagger}$, Yanbo Gao $^{2}$, Chuankun Li$^3$ and Wanqing Li$^4$}

\address{%
$^{1}$ \quad University of Electronic Science and Technology of China, Chengdu 611731, China; beidizhao@hotmail.com\\
$^{2}$ \quad Shandong University, Jinan 250061, China; shuaili@sdu.edu.cn; ybgao@sdu.edu.cn\\
$^{3}$ \quad North University of China, China; $^{4}$ University of Wollongong, Australia;}
\corres{Correspondence: shuaili@sdu.edu.cn}

\firstnote{Authors contributed equally to this work.}

\abstract{Smartphone sensors based human activity recognition is attracting increasing interests nowadays with the popularization of smartphones. With the high sampling rates of smartphone sensors, it is a highly long-range temporal recognition problem, especially with the large intra-class distances such as the smartphones carried at different locations such as in the bag or on the body, and the small inter-class distances such as taking train or subway. To address this problem, we propose a new framework of combining short-term spatial/frequency feature extraction and a long-term Independently Recurrent Neural Network (IndRNN) for activity recognition. Considering the periodic characteristics of the sensor data, short-term temporal features are first extracted in the spatial and frequency domains. Then the IndRNN, which is able to capture long-term patterns, is used to further obtain the long-term features for classification. In view of the large differences when the smartphone is carried at different locations, a group based location recognition is first developed to pinpoint the location of the smartphone. The Sussex-Huawei Locomotion (SHL) dataset from the SHL Challenge is used for evaluation. An earlier version of the proposed method has won the second place award in the SHL Challenge 2020 (the first place if not considering multiple models fusion approach). The proposed method is further improved in this paper and achieves 80.72$\%$ accuracy, better than the existing methods using a single model.}

\begin{document}


\section{Introduction}
Human activity recognition has been an active research area for decades and has lots of practical applications such as in video surveillance \cite{video surveillance1,video surveillance2,video surveillance3}, human-computer interaction \cite{human-computer interaction} and gaming \cite{gaming}. Nowadays, as the ubiquity, portability and the development of sensors in the mobile phone, there has been a growing interest in the research of smartphone sensors based human action recognition \cite{general0, general, general1, Smartphone-Based, Real-time}. Research on smartphone sensors based activity recognition for indoor localization \cite{Smartphone-Based}, real-time smartphone activity classification \cite{Real-time} and transportation recognition \cite{general1} has been actively investigated.

Different from the conventional video based human action recognition \cite{convention_video}, the data captured from smartphone sensors shows some specific characteristics. For example, due to the mechanism of smartphone sensors, it has been shown \cite{frontier} that the data are of periodic nature. Moreover, the sampling rate of the smartphone sensors is usually very high, resulting in a large number of very long-range data. Furthermore, different users have different living habits, and people usually place their mobile phones in different locations on their bodies, which causes large differences in the distribution of data. In addition to the large variance of data, the activity categories used in the smartphone sensors based classification are also different from the conventional human action recognition. Besides from the locomotion of a person, the transportation mode is also considered as an important classification task, including taking car, bus, train and subway, which could be very confusing.

To prompt the development of smartphone sensors based activity recognition, the Sussex-Huawei Locomotion (SHL) Challenge \cite{summary_2020} has been organized for three years from 2018 to 2020. It is based on the large-scale SHL dataset that was recorded over seven months by three participants engaging in eight transportation activities in real-life settings, including Still, Walk, Run, Bike, Car, Bus, Train and Subway {\cite{dataset1}}. This year's edition (2020) of the challenge \cite{summary_2020} aims to realize the user-independence and location-independence. 

There have been some works proposed in the literature for smartphone sensors based activity recognition, including conventional handcrafted features based and deep learning based methods. Especially with the rapid development of deep learning, lots of convolutional neural network (CNN) and recurrent neural network (RNN) based methods have been developed in the last few years. For the CNN based methods, EmbraceNet \cite{EmbraceNet} and DenseNet \cite{1D DenseNet} have been proposed for the task. However, due to the nature of convolution, its receptive field in the time domain is relatively small and the long-range temporal information cannot be well captured. On the other hand, due to the sequence processing capability of RNNs, RNNs are naturally appropriate for the task. In \cite{DCLSTMMWAR}, LSTM (long short-term memory) is used to process the sequence information. However, for the conventional RNNs including the simple RNN and LSTM, they usually suffer from the gradient vanishing and exploding problem, or gradient decay over layers due to the use of gates with non-saturated activation functions. Especially for the smartphone sensors based activity recognition, a model with long-range processing capability is highly desired. 

To address this long-range temporal processing problem, this paper developed a framework of combining short-term spatial/frequency feature extraction and long-term IndRNN recognition model. The contributions of this paper can be summarized as follows.

\begin{itemize}
\item A framework of combining short-term spatial and frequency domain feature extraction, and long-term IndRNN based recognition is proposed. The long-range temporal processing problem is divided into two problems to take advantage of the periodic characteristics of the sensor data.
\item A dense IndRNN model is developed to capture the long-term temporal information. Due to the capability of IndRNN in constructing deep networks and processing long-sequences, the dense IndRNN model can effectively process the short-term features to obtain the long-term information.
\end{itemize}

Preprocessing of derotating the sensor data to world coordinate system and postprocessing of transfer learning to new users in the test set are also used in the proposed method. Experimental results show that the proposed method achieves the state-of-the-art performance in the category of single-model based methods. An earlier version of the proposed method has appeared at a workshop paper for SHL Challenge 2020 \cite{No.2}. This paper further made a significant improvement by adding a detailed explanation of the proposed method and thorough analysis of the experiments with ablation study on the models and parameters. Moreover, feature augmentation with temporal changes is further developed, which improves the performance over the earlier one.

The rest of this paper is organized as follows. Section 2 describes the related work and the proposed method is presented and explained in Section 3. The experimental results and analyses are provided in Section 4. Section 5 concludes the paper.

\section{Related Work}
Vision based human activity recognition has been widely studied for decades, with many methods proposed in the literature. Such environmental sensors like cameras may become inconvenient in the open or crowed area to gather activity information of each individual. The distance between human and devices also affects the quality of signals, leading to differences in recognition accuracy. To address these issues, especially to collect the daily activity information on the basis of each individual in all areas, wearable sensors have become an attracting option. Some earlier wearable sensors, requiring markers on people, may also be intrusive and make people uncomfortable. However, with the quick popularization of the smartphone, smartphone sensors based human activity recognition is gaining interests since it does not require further devices other than the smartphone (most people already take them during the day). Many studies have been conducted for the activity recognition tasks based on smartphone sensors including recognizing the indoor activities \cite{Smartphone-Based}, the nursing activity to better care patients \cite{nurse}, also the movements people doing on their smartphones like typing and scrolling \cite{RTSACUISR}. Different approaches have been proposed for the smartphone sensors based activity recognition including the conventional handcrafted features based and the deep learning based methods, which will be briefly described in the following. 

\subsection{Conventional Handcrafted Features based Methods}
In the conventional handcrafted features based methods, spatial/temporal and frequency features are first extracted using techniques including statistical features such as mean, variance, standard deviation, maximum value, minimum value, energy, entropy and Fourier transform spectrums. Such features are engineered to capture the information over the sensor data. After features are extracted, some conventional machine learning methods such as Decision Trees \cite{ACURDWS}, KNN (k-nearest neighbors) \cite{UDOIHARMP}, Hidden Markov chain \cite{HMC} and SVM (support vector machine) \cite{SVM} can be used for the classification of the activity. In \cite{UDOIHARMP} and \cite{ACURDWS}, KNN and decision trees are used as classification models and the abovementioned spatial and frequency domain features are selectively used as input. In \cite{SVM1}, a 'one-versus-one' SVM is used to perform pairwise combinations selection and a Gaussian kernel is applied to process the features in a high dimensional space. In \cite{Random Forest+HMM}, random forest is used to predict the activity category of each frame first, then activities are smoothed over time with Hidden Markov chain considering that the activities in the daily life are continuous.

\subsection{Deep Learning based Methods}
\begin{figure}[!ht]
\centering
\includegraphics[width=12 cm]{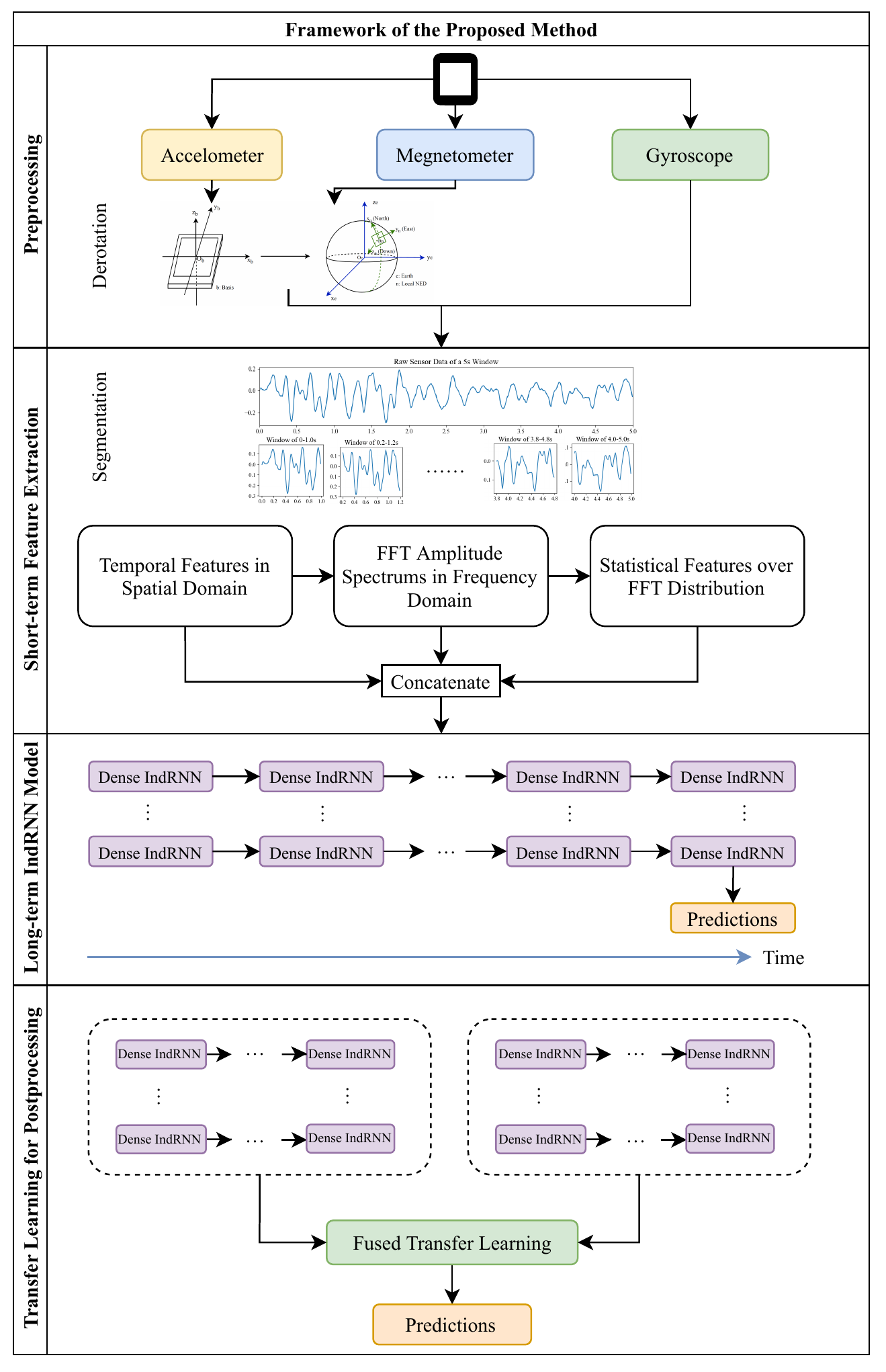}
\caption{Framework of the proposed method.}\label{fig_framework}
\end{figure}  
With the increasing applications and success of deep learning in many research areas, deep learning including both CNN and RNN has also been explored to perform the smartphone sensors based activity recognition. For the CNN based methods, Zeng et al. \cite{CNNHARUMS} and Zheng et al. \cite{TSCUMDCNN} used just one convolution layer as a spatial feature extractor to obtain the features at each time step, and then pooling is applied in the time direction to summarize the temporal information. However, with the shallow network and simple temporal processing technique, they cannot extract high-level spatial-temporal features and did not achieve a high accuracy. Charissa et al. \cite{HARSSUDLNN} employed a CNN using filters with a large time span 
to explore the long temporal correlation, and pooling over time is gradually used alternating with convolutional layers to reduce the loss over time. Zhu et al. \cite{1D DenseNet} proposed to use a 1D DenseNet model in order to take advantage of deeper CNNs. The DenseNet is first applied on each sensor independently and then combined together. All the data in the time domain are sampled and provided as one input to network to better explore the temporal information. Considering the large volume of the temporal data, this also results in a large number of parameters. Choi et al. proposed an EmbraceNet \cite{EmbraceNet} to fuse multiple CNN models together. It also processes each sensor independently and then combine them together. In all, the CNN based methods usually process the temporal sequence with pooling or convolution, which is not very effective in the long-range problem.

\begin{figure}[!ht]
\centering
\includegraphics[width=15cm]{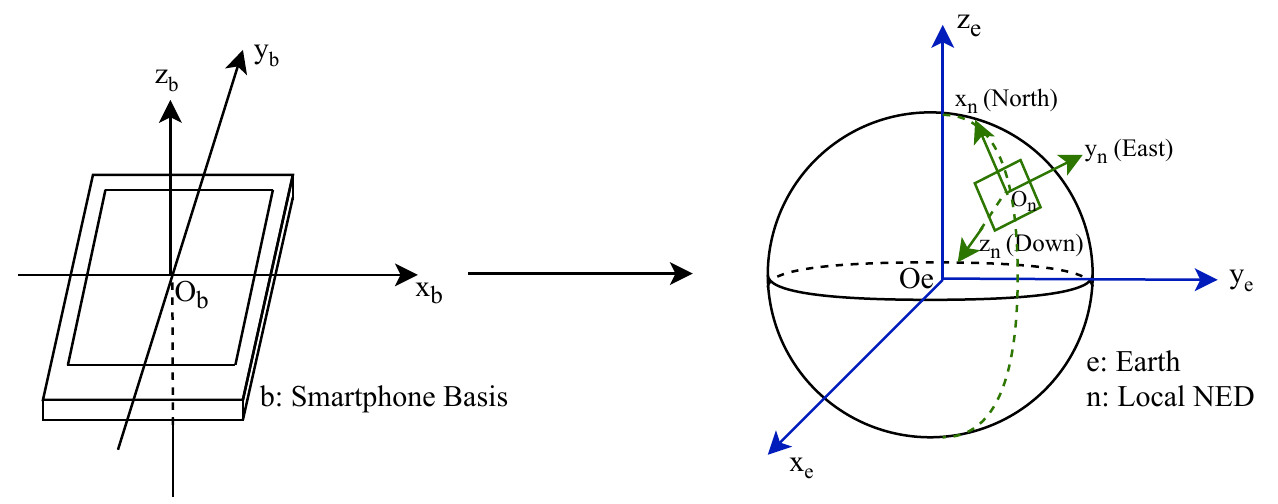}
\caption{Derotation of coordinates from the smartphone coordinate system to the NED (North-East-Down) coordinate system.}\label{fig_derotation}
\end{figure}  
Since the smartphone sensors based human activity recognition is a temporal sequence processing task, RNN can be naturally selected with its temporal processing capability. Francisco et al. proposed a deep framework \cite{DCLSTMMWAR} using convolution and LSTM (long short-term memory) together where the convolution extracts the spatial feature and LSTM helps learn the long-term temporal information. However, the gate mechanism in LSTM makes it difficult to construct deep networks. Some researchers migrate the dense and residual architecture to LSTM to assist constructing deep networks, but the performance of improvement is not significant \cite{deepLSTM}. In \cite{RNN1}, Rui et al. first used a dilated convolutional neural networks to extract local short-term features. Then a shallow dilatedSRU is developed to model the long temporal dependencies. In a word, the conventional RNN models used for classification are usually shallow and cannot effectively construct deep models due to the gradient decay within each layer. On the contrary, the recently proposed IndRNN \cite{IndRNN2018, IndRNN2019} has been show to be able to better explore the high-level and long-term information, which has also been used in the last two years' SHL Challenge \cite{IndRNN_for_SHL_2018, IndRNN_for_SHL_2019} as the base module with only the spatial information or FFT magnitudes using a relatively shallow network. This paper further proposes a framework of combining short-term spatial and frequency features, and long-term deep dense IndRNN models for activity recognition.

\section{Proposed Method}

\subsection{Overall Framework}
This paper proposed an Independently Recurrent Neural Network based long-term activity recognition method based on short-term spatial and frequency domain features. The framework of the proposed method consists of four modules as shown in Fig. \ref{fig_framework}: preprocessing, short-term spatial and frequency feature extraction, long-term IndRNN model and transfer learning for postprocessing. Among them, the preprocessing and short-term feature extraction modules process the input data to short-range spatial features and frequency domain features to accommodate the periodic nature of the smartphone sensors data. Then the IndRNN model, taking advantage of its capability of processing very long sequences and constructing deep models, is applied as the main recognition model to solve the long-range classification problem. Finally, transfer learning is adopted as postprocessing to fine-tune the model in order to realize user-idependence. Details on each module are presented in the following.

\subsection{Preprocessing}
\begin{figure}[!ht]
\centering
\includegraphics[width=15cm]{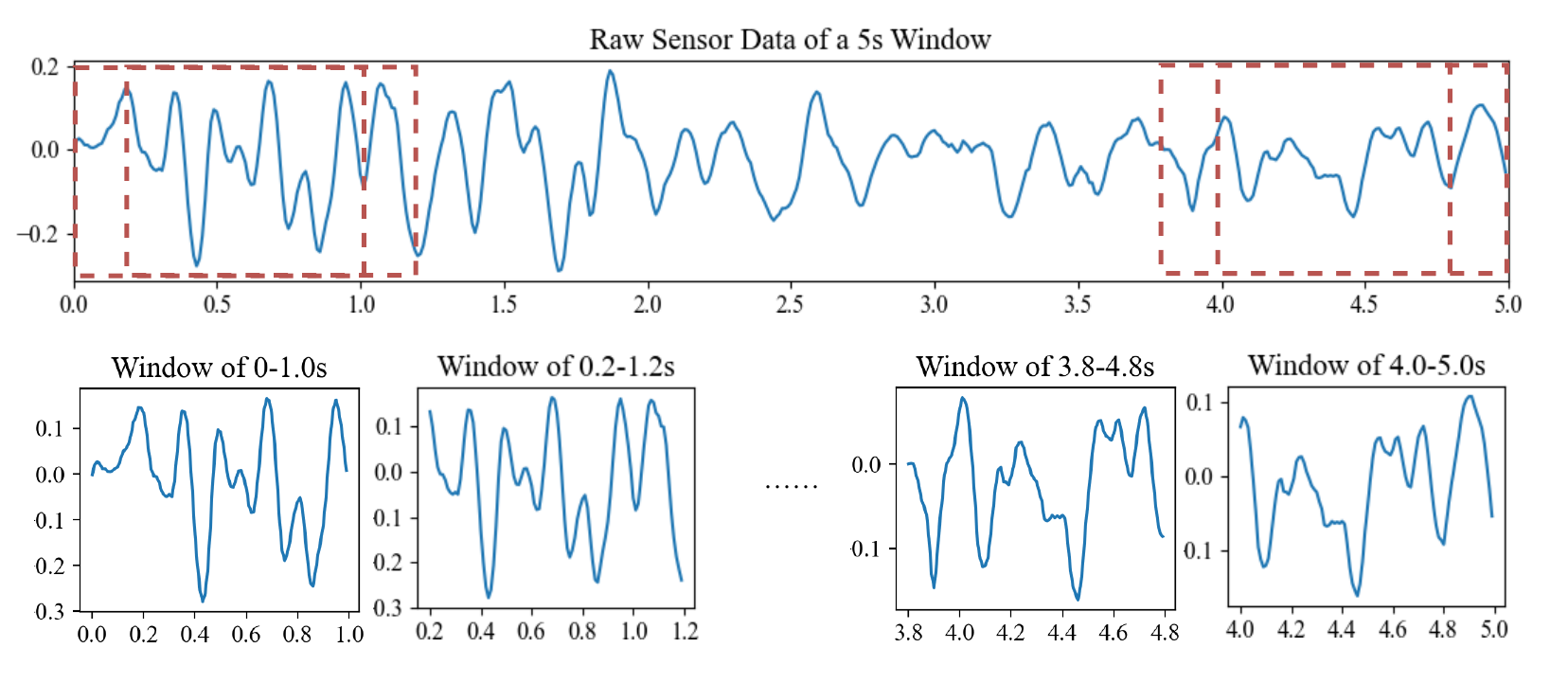}
\caption{Illustration of the short-term data segmentation.}\label{fig_seg}
\end{figure}  

For the current smartphones such as HUAWEI Mate 9 used to collect data in the SHL dataset \cite{dataset1,dataset2}, the sensor data is measured in a coordinate according to the smartphone position. The basis of triaxial sensors is $(x_{b}, y_{b}, z_{b})$ where, for most phone, $x_{b}$ is along the shorter side and pointing right, $y_{b}$ is along the longer side and pointing up and $z_{b}$ is perpendicular to the screen and pointing out. The accelerometer and magnetometer sensors, two of the smartphone sensors, measure the acceleration of the device and the magnetic field of the earth at the device location, respectively. They are represented by two 3-dimensional vectors, which represent the acceleration of the phone and the magnetic field of where the phone is, respectively. Since the data is measured in the coordinate according to the smartphone position, the sensor data can be inconsistent in the world coordinate when only the phone is rotating without the user's body movement. In turn, it will affect the classification accuracy of the user's activity without preprocessing. Therefore, to reflect the real movement of the user in the world coordinate, the sensor data needs to be derotated to the consistent world coordinate system.

In this paper, the NED (North-East-Down) coordinate system is used to transform the sensor data as shown in Fig. \ref{fig_derotation}, where $x_{n}$ points toward East, $y_{n}$ points toward magnetic North and $z_{n}$ points up toward the sky. The transform can be performed by multiplying the raw sensor data with the rotation matrix $R$ derived from orientation sensor of the device in quaternions $[q_w,q_x,q_y,q_z]$ as shown in equation (\ref{equation1}) and (\ref{equation2}).

\begin{equation}\label{equation1}
R=\left[\begin{array}{ccc}
1-2\left(q_{y}^{2}+q_{z}^{2}\right) & 2\left(q_{x} q_{y}-q_{w} q_{z}\right) & 2\left(q_{x} q_{z}+q_{w} q_{y}\right) \\
2\left(q_{x} q_{y}+q_{w} q_{z}\right) & 1-2\left(q_{x}^{2}+q_{z}^{2}\right) & 2\left(q_{y} q_{z}-q_{w} q_{x}\right) \\
2\left(q_{x} q_{z}-q_{w} q_{v}\right) & 2\left(q_{v} q_{z}+q_{w} q_{x}\right) & 1-2\left(q_{x}^{2}+q_{v}^{2}\right)
\end{array}\right]
\end{equation}
\begin{equation}\label{equation2}
\left[\begin{array}{l}
x_{n} \\
y_{n} \\
z_{n}
\end{array}\right]=R\left[\begin{array}{l}
x_{b} \\
y_{b} \\
z_{b}
\end{array}\right]
\end{equation}
where $(x_{n},y_{n},z_{n})$ represents the transformed data in the NED coordinate system, which is consistent to the user's movement. The transformed data can then be used for the following feature extraction.

\subsection{Short-term Spatial and Frequency Domain Feature Extraction}
\begin{figure}[!ht]
\centering
\includegraphics[width=15.5cm]{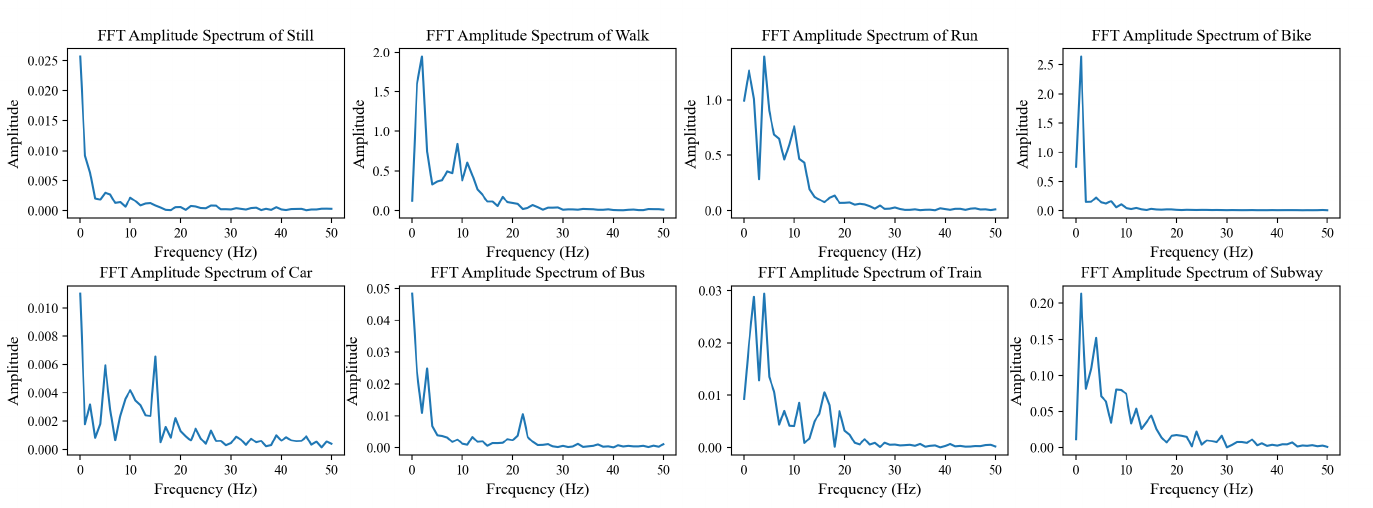}
\caption{Example FFT amplitude spectrums from one segmented window of different classes.}\label{fig_freq}
\end{figure}  

For the sensors used in the HUAWEI Mate 9, the sampling rate is 100Hz. Data from a window of 5 seconds is used for each classification, resulting in 500 frames of data. Generally, processing of a very long-range data such as 500 steps is difficult due to the complex temporal pattern. Also the data from the smartphone sensors has been shown to be periodic \cite{frontier}. Therefore, some short-term spatial and frequency domain features are extracted first as explained following. 

First, the data of each 500-frames (5 seconds) sample was segmented into 21 100-frames (1 second) overlapping sliding windows as shown in Figure \ref{fig_seg}. Each segmented window contains short-term signals and long-time signals can be obtained by combining them over time. The data from 7 sensors are provided for classification, including accelerometer, gyroscope, magnetometer, linear acceleration, gravity, orientation, and ambient pressure, resulted in a total of 20 channels of data. Since accelerometer is a superposition of the linear acceleration and gravity, the linear acceleration and gravity data are not used in order to reduce the size of the data input. Also since orientation is used to derotate the other sensors data, it is no longer used after the preprocessing. In all, the data from gyroscope, derotated data from accelerometer and magnetometer, and pressure are used in our method, which contains 10 channels.

For each segmented window, some spatial features over time are first extracted including mean, numbers above mean, numbers below mean, standard deviation, minimum value, maximum value similarly as in \cite{RTSACUISR}. And for pressure, the data is normalized per sample and used as input to show the change within each sample. The pressure data performs not well in activity recognition, but well in location recognition model introduced later. The description of the features are shown in Table \ref{tab_spafea}. On the other hand, due to the strong periodicity of the smartphone sensor data, fast Fourier transform (FFT) is used to transform the data into the frequency domain. The FFT amplitude spectrums are then extracted as features where only the magnitudes of the coefficients are used (half of the total data). Some examples of the FFT amplitude spectrums from all the classes are shown in Fig. \ref{fig_freq}. It can be seen that the distribution of FFT amplitude spectrums can be quite different among different classes. Therefore, in addition to the amplitude spectrum, some statistical features on top of the frequency features including mean and standard variation are also extracted and combined with previous features.


\begin{table}[!ht]
\caption{Extracted short-term features in the spatial-temporal domain and their definitions.}\label{tab_spafea}
\centering
\begin{tabular}{ccc}
\toprule
\textbf{Time Domain Features}	& \textbf{Description}	\\
\midrule
Mean& The average value of the data for each axis in the window\\
Numbers above Mean&The numbers of values above the mean of the window	\\
Numbers below Mean&The numbers of values below the mean of the window	\\
Standard Deviation&Standard deviation of each axis in the window	\\
Minimum Value& The minimum value of the data for each axis in the window	\\
Maximum Value&The maximum value of the data for each axis in the window	\\
Per Sample Normalized Pressure&The normalized pressure of each sample	\\
\bottomrule
\end{tabular}
\end{table}


\subsection{Long-term IndRNN (Independently Recurrent Neural Network) Model}
With the short-term spatial/temporal and frequency domain features extracted, a long-term recognition model is further proposed for the final recognition. In this paper, our previously proposed Independently Recurrent Neural Network (IndRNN) \cite{IndRNN2018, IndRNN2019} is adopted as the basic model. The structure of the IndRNN \cite{IndRNN2018, IndRNN2019} follows

\begin{equation}
\boldsymbol{h}_{t}=\sigma\left(\boldsymbol{W} \boldsymbol{x}_{t}+\boldsymbol{u} \odot \boldsymbol{h}_{t-1}+\boldsymbol{b}\right)
\end{equation}
where $\boldsymbol{x}_{t} \in R^{M}$ and $\boldsymbol{h}_{t} \in R^{N}$ is the input and hidden state at time step $\mathrm{t},$ respectively. $\boldsymbol{W} \in R^{M \times N}, \boldsymbol{u} \in R^{N}$ and $\boldsymbol{b} \in R^{N}$ are the weights for the current input and the recurrent input and the bias of neurons. $\odot$ represents the Hadamard product and $\sigma$ is the nonlinear activation function of neurons. $\mathrm{N}$ is the number of neuron of this IndRNN layer. With this form, each neuron in IndRNN is independent from each other and the gradient backpropagation can be calculated for each other. Accordingly, by regulating the recurrent weights, it well addresses the gradient vanishing and exploding problems. Therefore, it can process very long sequences. Also it can work very robustly with non-saturated functions such as ReLU, thus is able to construct very deep networks. 
\begin{figure}[!ht]
\centering
\subfigure[The structure of dense IndRNN layer and dense IndRNN block.]{
\begin{minipage}[t]{1\linewidth}\label{structure_a}
\centering
\includegraphics[width=10.5cm]{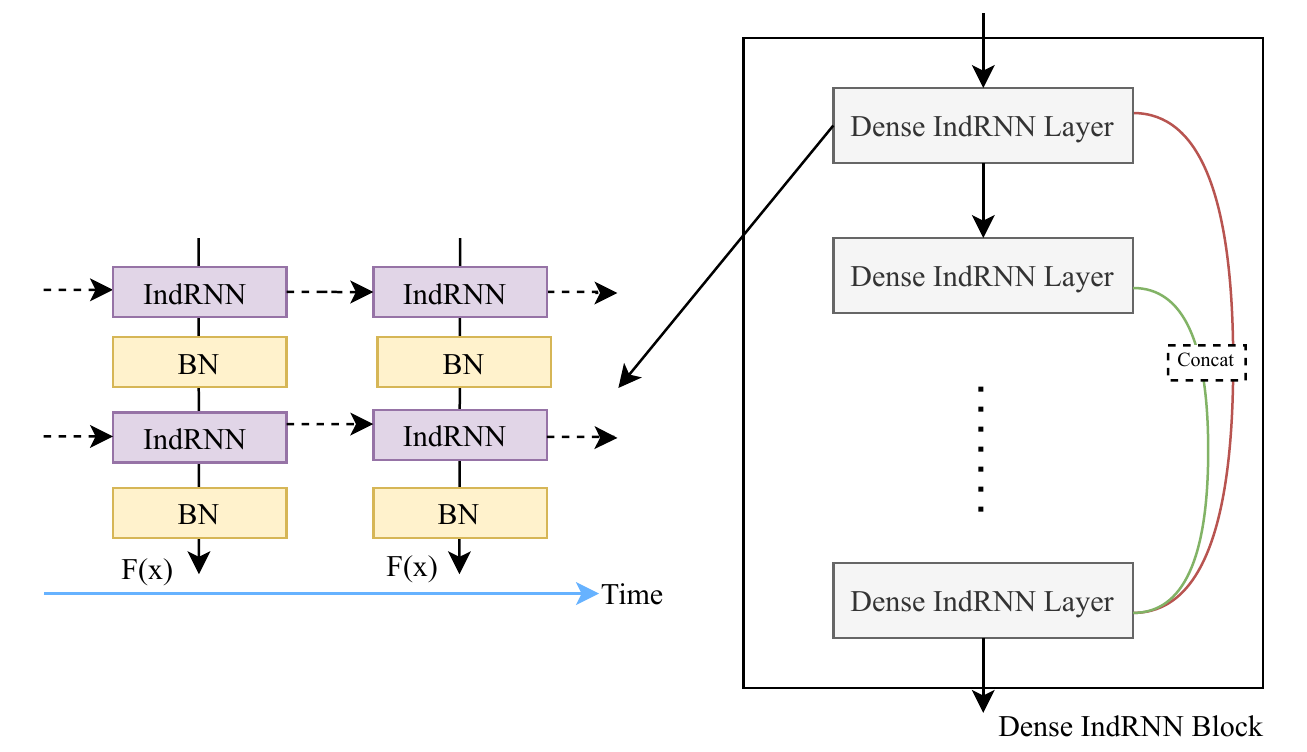}
\end{minipage}%
}%
\quad
\subfigure[The architecture of the dense IndRNN model.]{
\begin{minipage}[t]{1\linewidth}\label{structure_b}
\centering
\includegraphics[width=12.5cm]{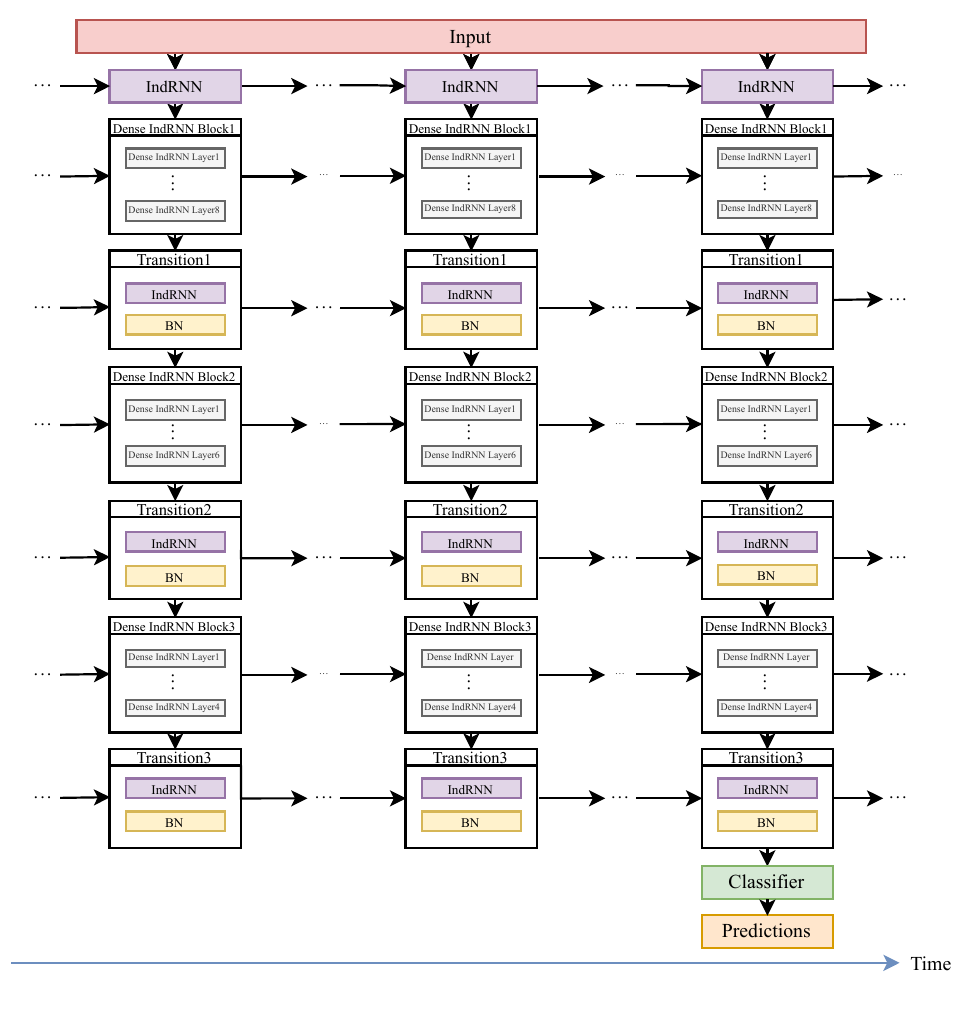}
\end{minipage}%
}%

\centering
\caption{Illustration of the proposed dense IndRNN structure.}
\end{figure}

\begin{figure}[!ht]
\centering
\subfigure[Confusion matrix of the location recognition on the validation set: on four locations.]{
\begin{minipage}[!ht]{0.45\linewidth}
\centering
\includegraphics[width=3in]{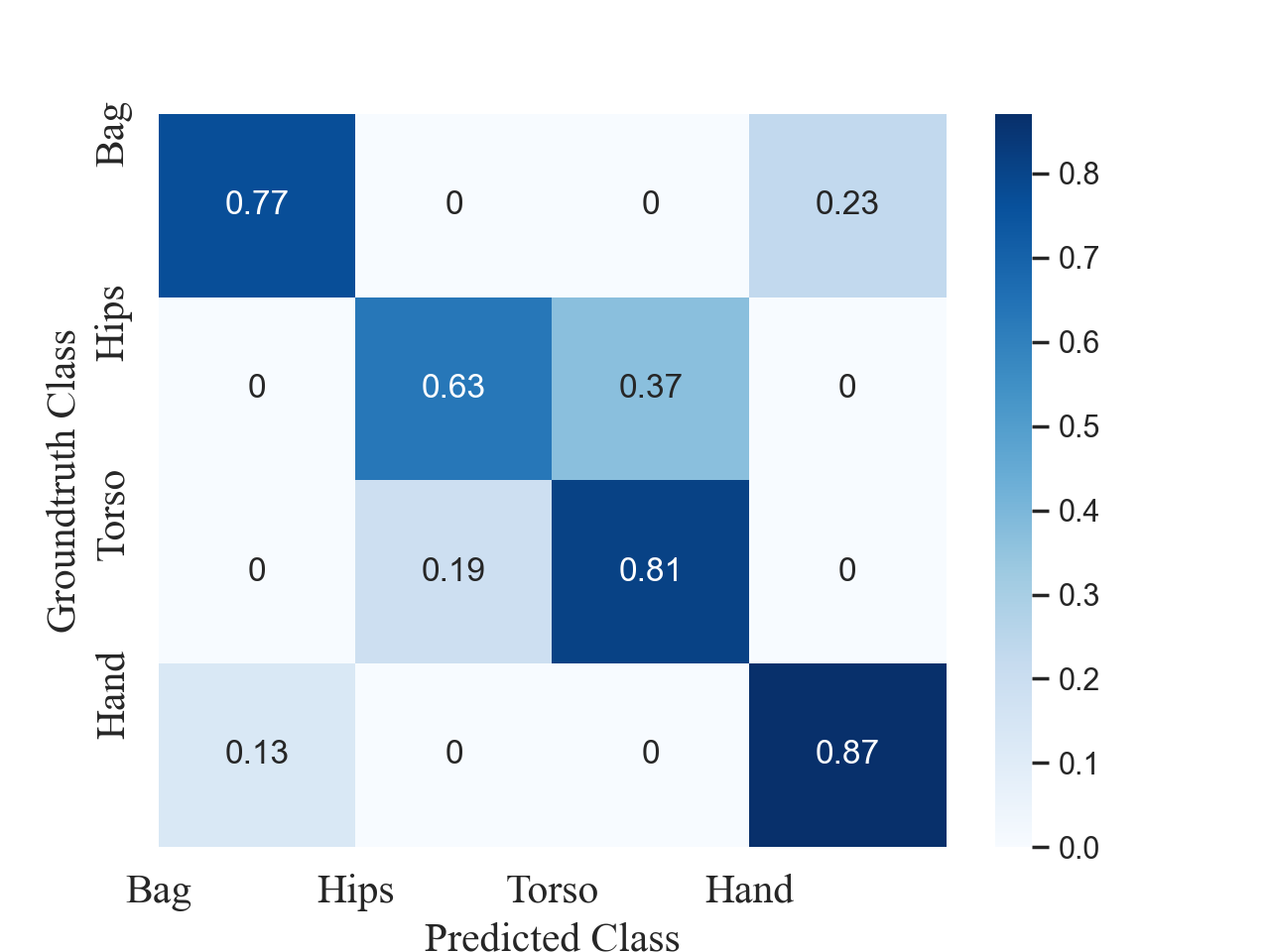}\label{loc1}
\vspace{0.01cm}
\end{minipage}%
}%
\hfill
\subfigure[Confusion matrix of the location recognition on the validation set: two groups - Bag and Hand, Hips and Torso.]{
\begin{minipage}[!ht]{0.45\linewidth}
\centering
\includegraphics[width=3in]{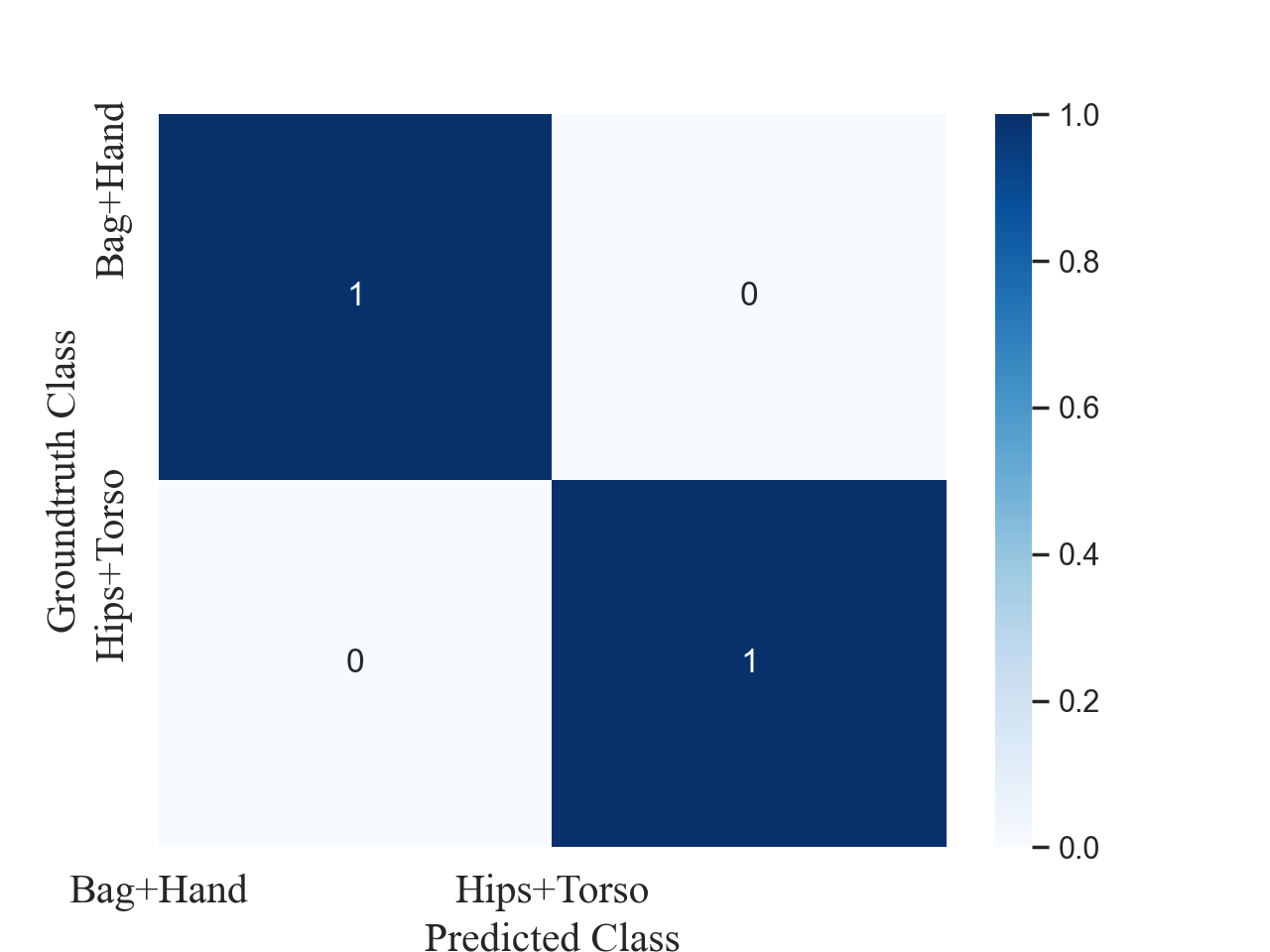}\label{loc2}
\vspace{0.01cm}
\end{minipage}%
}%
\centering
\caption {Confusion matrices of the location recognition on the validation set.}\label{fig_location}
\end{figure}

In this paper, we propose to use a deep dense IndRNN as the main classification model. The diagram of the proposed dense IndRNN model is shown in Fig. \ref{structure_b} and the detailed illustration of each dense layer and dense block is shown in Fig. \ref{structure_a}. The overall architecture follows \cite{IndRNN2019}. It consists of three dense blocks with 8, 6, 4 dense layers, and each dense layer contains two IndRNNs as shown in Fig. \ref{structure_b}. Batch normalization is used after each IndRNN layer to accelerate training. Dense architecture concatenates feature output from all the previous dense layers in a dense block as the input for the next dense layer. It facilitates the feature reuse of the relatively shallow layers. After each dense block, a transition block with one IndRNN layer is followed to compress the features as a bottleneck, where the outputs are usually reduced to the half of the input features. At last, a classifier with one linear function and softmax activation is used at the last time step for the final classification.

The cross-entropy loss is used as the objective function for training, which is
\begin{equation}
L=\sum_{i=1}^{8} t_{i} \log \left(p_{i}\right)
\end{equation}
where $t_{i}$ is an indicator variable, which equals to 1 when the prediction is right, 0 when the prediction is wrong. $p_{i}$ is the predicted probability of this sample. The categorical cross-entropy has been widely used for classification.

\subsection{Transfer Learning for Post-processing}\label{subsec_trans}
With the above preprocessing, short-term feature extraction and long-term IndRNN based recognition, different activities can be classified. However, considering that the smartphone can be placed at any place by the user such as holding in the hand, bag, or in the lap pocket, the sensor data can be of large differences. Directly classifying different sensor data captured from different locations can be difficult, and the most appropriate features used for classification under different locations may also be different. Therefore, in view of the differences among different sensors, the location of the sensor data is first recognized. Then in the test, we can pinpoint the location of the data and use an appropriate model for classification. In this process, the labels of the sensor data are changed to the locations of the sensors. A simple plain IndRNN model of stacking 6-layer IndRNNs is used for the classification. 

The location recognition result in terms of the confusion matrix is shown in Fig. \ref{loc1}, where four locations are used, including bag, hips, torso and hand. It can be observed that while different locations can be recognized with a relatively good accuracy, there are still some confusion among different classes, especially between bag and hand, hips and torso. If locations are recognized into two groups, bag and hand, hips and torso, the classification of two groups can be very accurate as shown in Fig. \ref{loc2}. It indicates that the features of the data from each group can be similar while the features from different groups can be very distinguished. Therefore, in the proposed scheme, a group based location recognition is used, where the data is first classified into two groups and then further recognized as different activities. Note that in the SHL dataset used in the experiment, all the data from the test set comes from one unknown location, thus is classified first to one location group and only one model is constructed for this recognized location group. 

\begin{figure}[!ht]
\centering
\subfigure[Distribution of user2's labels over the validation set.]{
\begin{minipage}[t]{0.5\linewidth}
\centering
\includegraphics[width=3in]{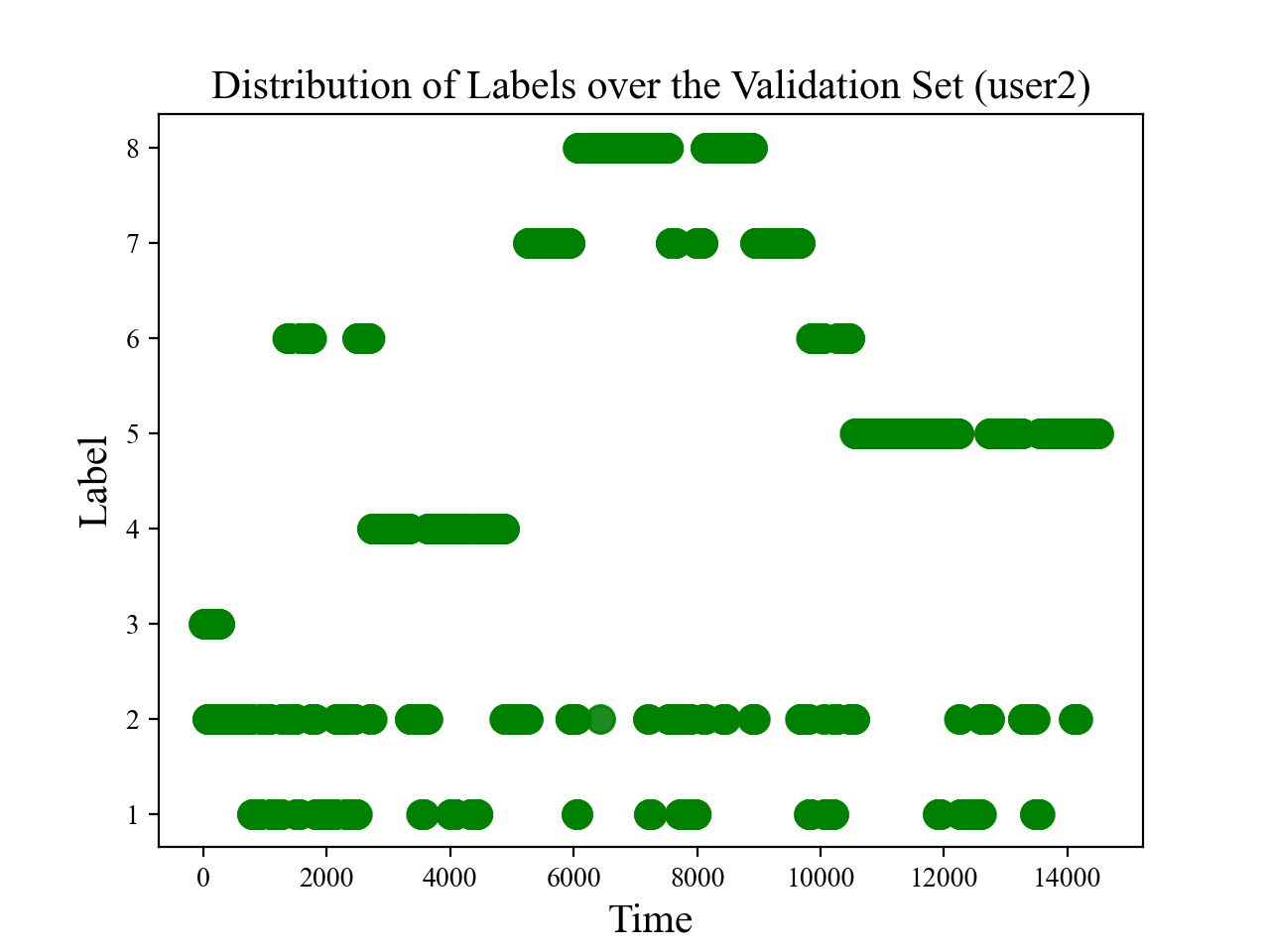}
\end{minipage}%
}%
\subfigure[Distribution of user3's labels over the validation set.]{
\begin{minipage}[t]{0.5\linewidth}
\centering
\includegraphics[width=3in]{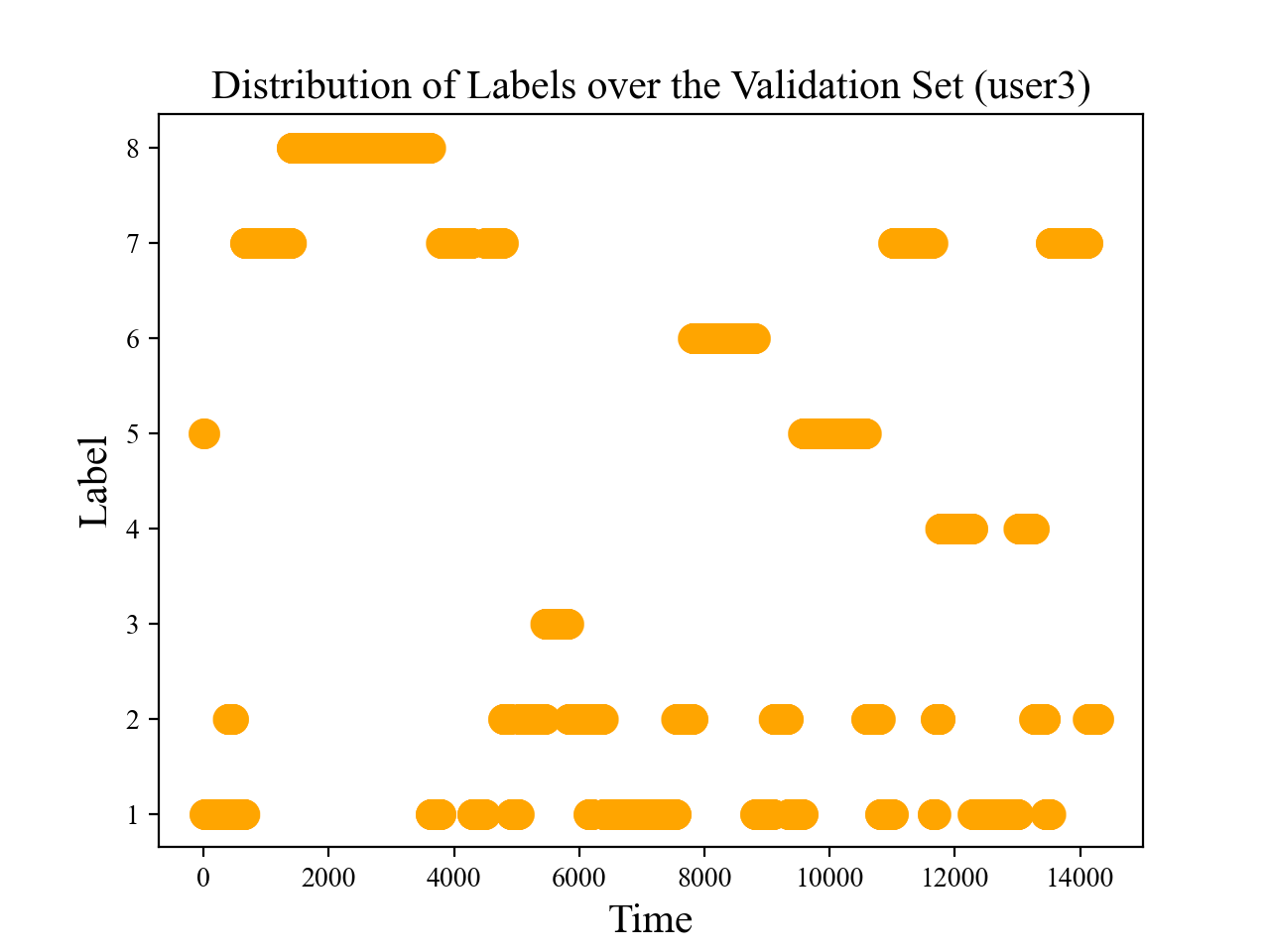}
\end{minipage}%
}%
\centering
\caption{Distribution of labels over the validation set.}\label{fig_valdis}
\end{figure}

On the other hand, due to limitation of the dataset which only contains data from three users (although with a large amount of data - 196072 frames), transfer learning is used to quickly generalize the model to different users. In the SHL dataset, only user1 is used as training data, a small amount of data from the other two users are used as validation data, and the remaining data from the user2 and user3 are kept for test. To fully take advantage of the validation data (which is allowed in the challenge), the validation data is first split and part of it is used to transfer the model learned on the training data of user1 to the test data of user2 and user3. For simplicity, the learned model is directly fine-tuned on the transfer data. The most common way of transfer learning is to use a half of the validation set as transfer training set and another half acts as transfer validation set. However, in this challenge, splitting the validation set directly into parts may lead to over-fitting because labels of the validation set distribute unevenly as shown in Figure \ref{fig_valdis}. Therefore, the data with same labels is first stacked together, then divided with a similar proportion of data from all the classes to construct the transfer training set and the transfer validation set for the transfer learning process.

\begin{figure}[!ht]
\centering
\includegraphics[width=15 cm]{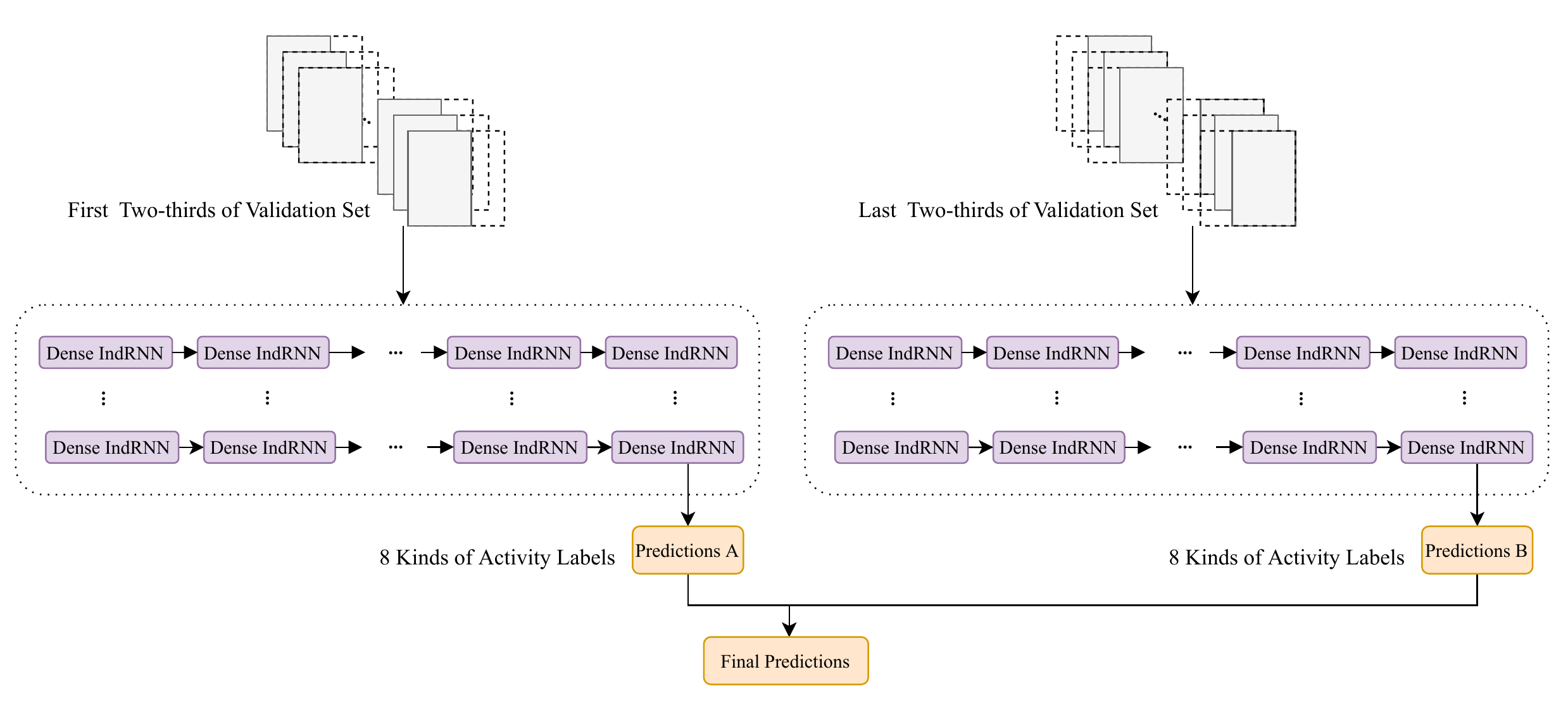}
\caption{Diagram of the fused transfer learning.}\label{fig_transfer}
\end{figure}

When conducting the transfer learning process, it leads to different accuracies using the first half and the second half of the original validation set for training because of the limited size of the validation set. Accordingly, we further swap the transfer training and transfer validation set to learn two models, noted as TransferA and TransferB, and then fuse them together to take advantage of all the data. The diagram of the transfer learning is shown in Fig. \ref{fig_transfer}.

\section{Experimental Results}

\subsection{Dataset and Setup}
\subsubsection{SHL Dataset}
SHL dataset \cite{dataset1}\cite{dataset2} is used for evaluation in this paper, which is also the dataset used in the SHL Challenge 2020. It was recorded over seven months in 2017 from three users (user1, user2 and user3). The aim of this dataset is to use machine learning methods and heuristics to realize the recognition of users’ 8 locomotion modes and transportation (Still, Walk, Run, Bike, Bus, Car, Train and Subway). The smartphone used to collect data is put on four locations on the body (Bag, Hips, Torso and Hand). The dataset aims to realize the user-independence and location independence. To be specific, the training set contains $272\times4$ hours from four locations of user 1. The validation set consists of $40\times4$ hours data from four locations of the combination of user2 and user3. The test set contains 160 hours data of user2 and user3 from an unknown location (Hips after the Challenge result is published). 

The data is collected from 7 raw sensors, including accelerometer, magnetometer, gyroscope, magnetometer, linear acceleration, gravity, orientation, and ambient pressure, which combines a total of 20 channels. The sampling rate is 100Hz, and all of the data is segmented into 5 seconds windows. So the data size of the training set, validation set and test set are $196072\times500$, $28789\times500$ and $57573\times500$, respectively.

\subsubsection{Training Setup}
For training, Adam \cite{Adam} is used for optimization. The learning rate of our model is set to $2\times10^{-4}$ at first. To restrain the slightly larger fluctuation at the beginning of the training process, it is set to $2\times10^{-5}$ at the first 10 epochs as a learning rate warmup strategy. The learning rate drops 10 times once the validation accuracy does not increase (over a large patience 100). Mini-batch with 128 batch size is used to train our model. The dense block configuration is set to (8, 6, 4), where in the first, second and third dense block, 8, 6 and 4 dense layers are used, respectively. This keeps a relatively similar number of neurons in each dense block. The growth rate is set to 48. 

In our model, ReLU is applied as an activation function. Compared to the tanh and sigmoid function, it not only reduces the amount of computation but also helps to alleviate the problem of gradient vanishing. In order to reduce over-fitting, dropout is applied after the input (0.5), each dense layer (0.5), each bottleneck layer (0.1) and each transition layer (0.3).

\subsubsection{Evaluation}

\begin{figure}[!ht]
\centering

\subfigure[Confusion matrix of plain IndRNN.]{
\begin{minipage}[t]{0.45\linewidth}
\centering
\includegraphics[width=3in]{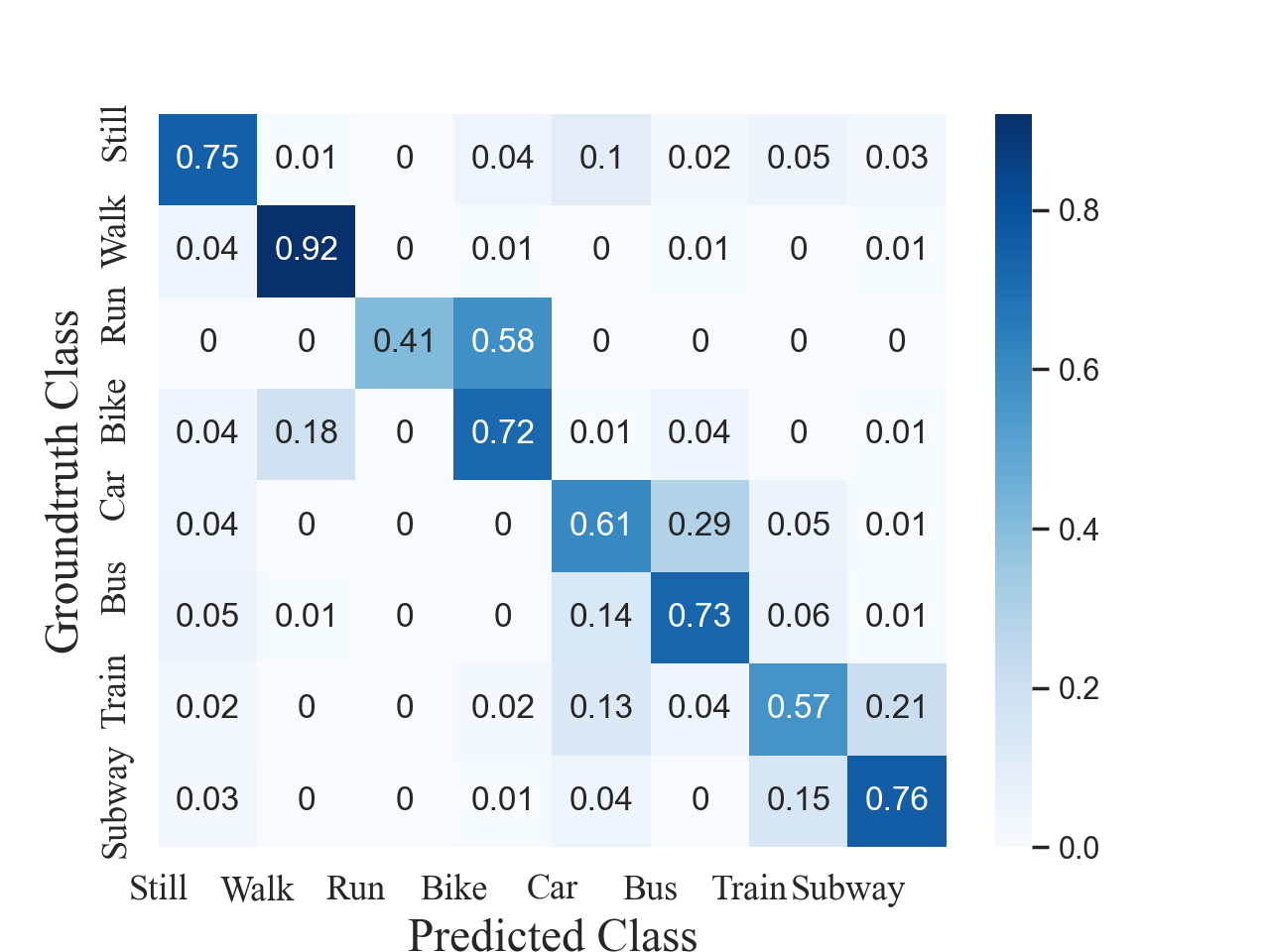}
\end{minipage}%
}%
\subfigure[Confusion matrix of residual IndRNN.]{
\begin{minipage}[t]{0.45\linewidth}
\centering
\includegraphics[width=3in]{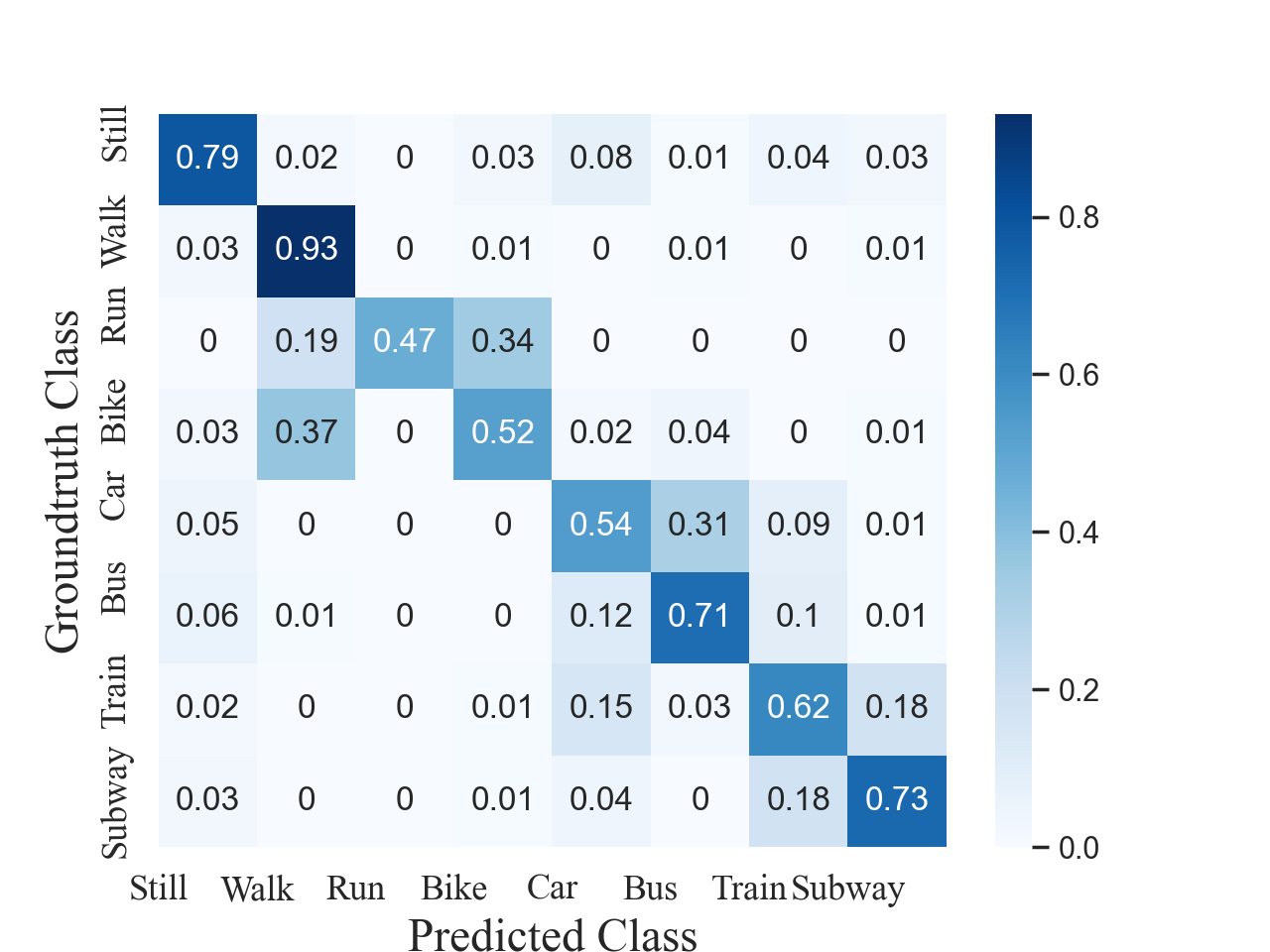}
\end{minipage}%
}%

\subfigure[Confusion matrix of dense IndRNN.]{
\begin{minipage}[t]{0.45\linewidth}
\centering
\includegraphics[width=3in]{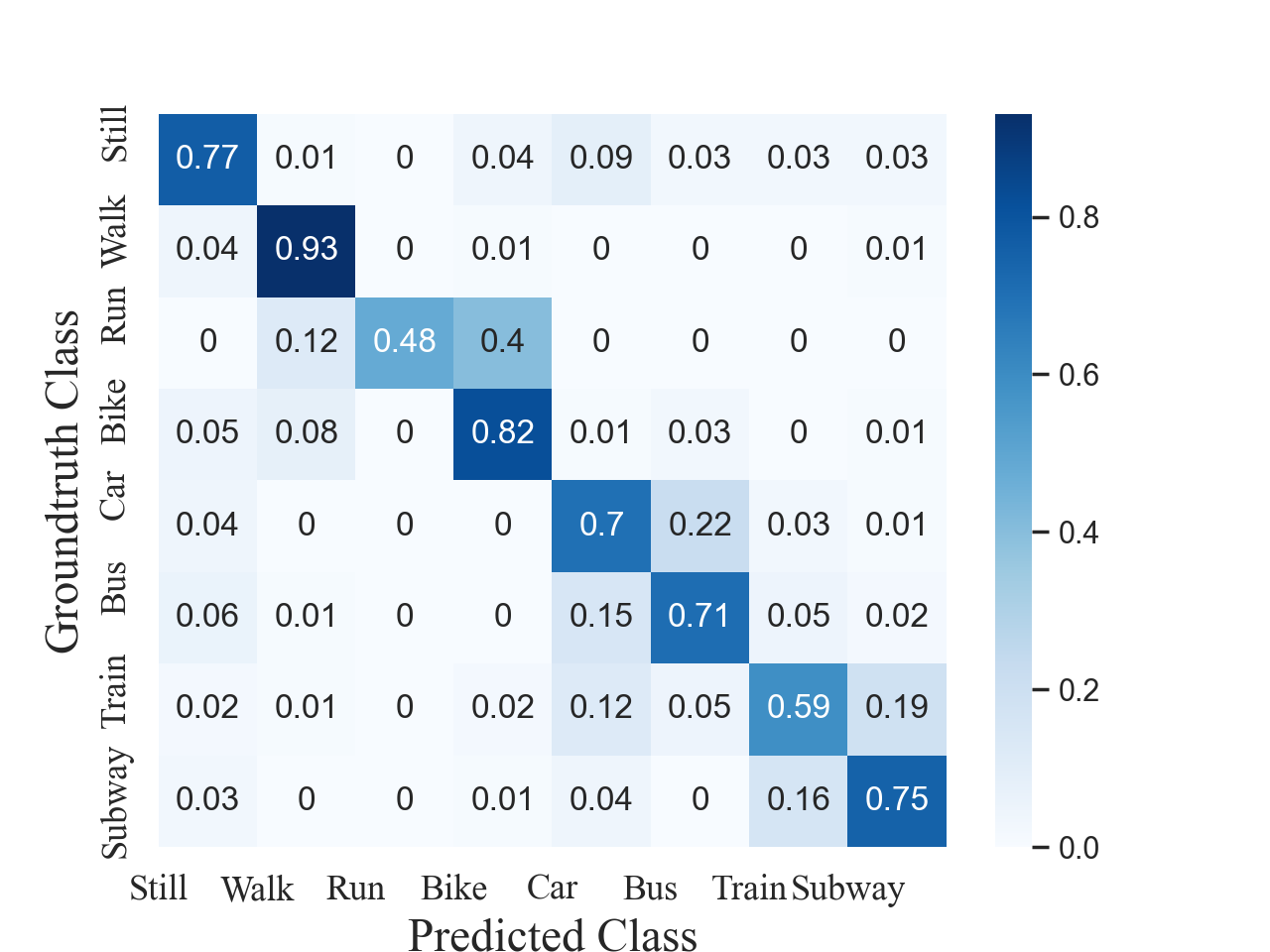}
\end{minipage}
}%
\subfigure[Confusion matrix of dense-aug IndRNN.]{
\begin{minipage}[t]{0.45\linewidth}
\centering
\includegraphics[width=3in]{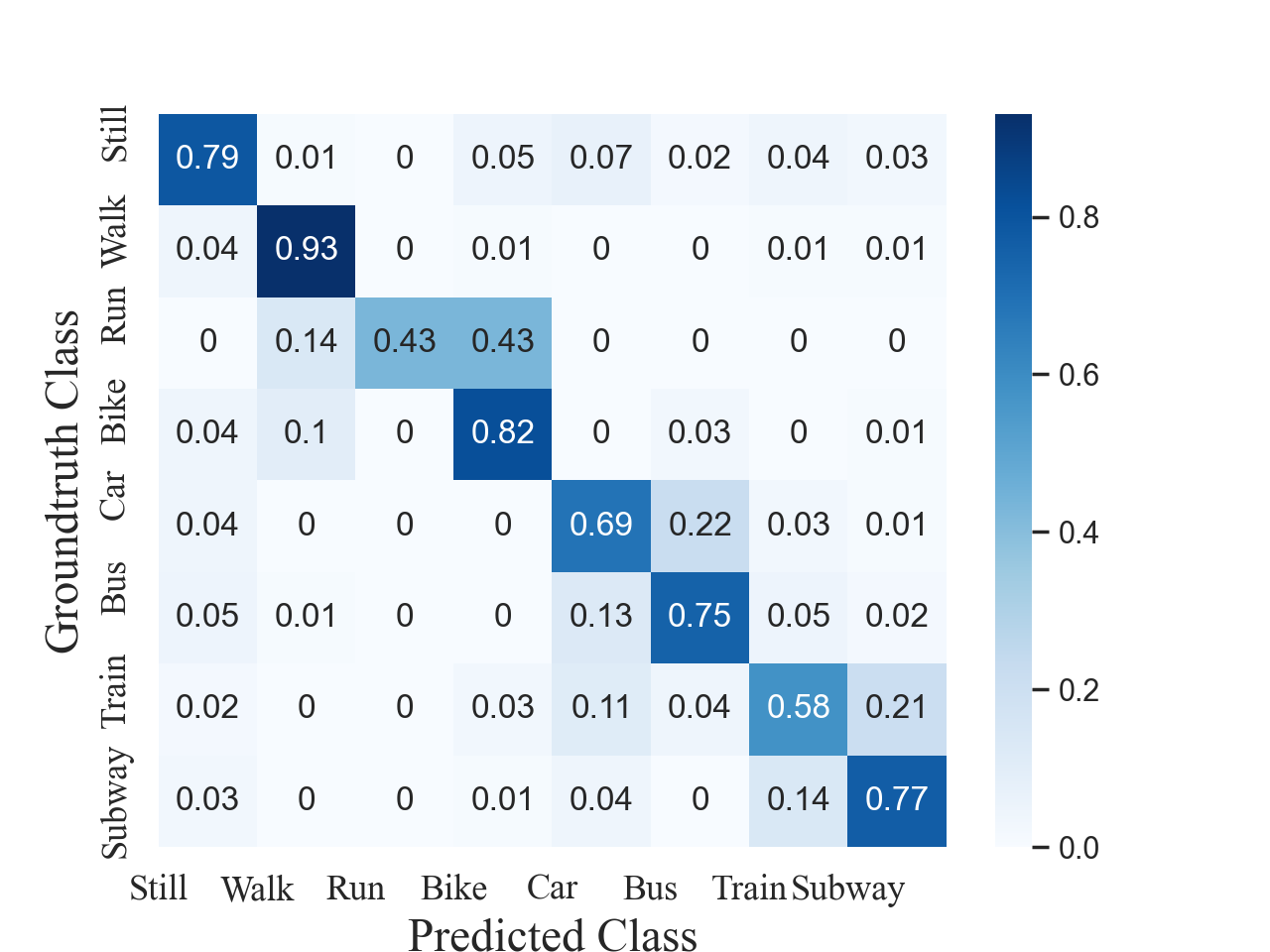}
\end{minipage}
}%
\centering

\caption{Confusion matrices of different IndRNN architectures.}\label{Confusion of four IndRNN}
\end{figure}
The final performance is evaluated using the F1 score. It can better reflect the performance when the distribution of the data is imbalanced among different classes. Traditionally, the F1 score is used in evaluating binary classifications and can be defined with precision and recall as follows:
\begin{equation}
\begin{array}{l}
\text {Precision}=\frac{T P}{T P+F P} \\
\\
\text {Recall}=\frac{T P}{T P+F N} \\
\end{array}
\end{equation}
where $TP$ represents True Positive, $TN$ is True Negative, $FP$ is False Positive and $FN$ is False Negative. Among them, Precision focuses on assessing how much of all the data that is predicted to be positive is the true positive. Recall focuses on how many samples are successfully predicted to be positive among those real positive. In multi-categories classification, the precision and recall are calculated for each class separately, and the overall precision, recall, F1-score can be obtained as follows:

\begin{equation}
\begin{array}{l}
	\text{Precision}=\frac{P_{still}+…+P_{subway}}{8} \\
	\\
	 \text{Recall}=\frac{R_{still}+…+R_{subway}}{8}\\
	 \\
	\text{ F1-score}=\frac{2 \times  \text{ Precision } \times \text{ Recall }}{\text{ Precision }+\text{ Recall }}
\end{array}
\end{equation}
	
In the SHL Challenge, since the location is unknown, location recognition is first performed to recognize the location of the test set. In this paper, since the location is already reported, the validation data from the known location (Hips) is used for validation. It is observed that there is no large difference using a group based location or a specific location. Also in the practical applications, we argue that the locations are always unknown and the group based location may better describe the data as shown in Fig. \ref{fig_valdis}.

\subsection{Ablation Studies on Models, Augmentation and Learning Rates}
Firstly, three different model architectures are evaluated including the plain IndRNN, residual IndRNN and dense IndRNN. The results on the test set are shown in Table \ref{ablation_model} and the confusion matrices are shown in Fig. \ref{Confusion of four IndRNN}. It can be seen that the dense IndRNN performs the best. Therefore, in the following experiments, dense IndRNN is used as the baseline of the model. 

On the other hand, feature augmentation is also explored in the proposed method. In addition to the input data and features at each time step for input of the network and deeper layers of the network, this paper also augments the input data and features with the temporal difference. The augmentation can be viewed as a form of optical flow in the video based classification tasks. It provides the first-order change information for better processing. The result is also shown in Table \ref{ablation_model}, and it can be seen that the feature augmentation improves the performance.

\begin{figure}[t]
\centering
\includegraphics[width=14 cm]{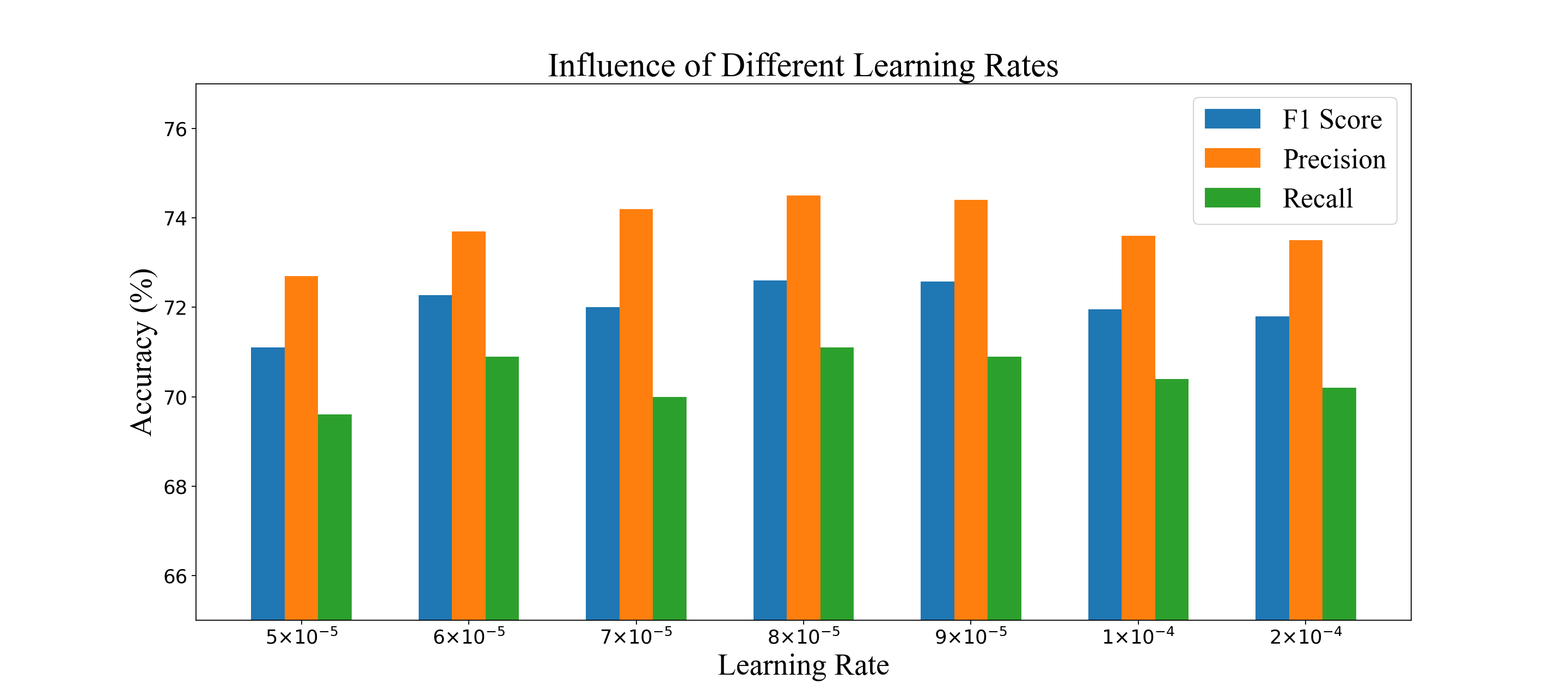}
\caption{Illustration of using different learning rates.}\label{lr}
\end{figure}  
\begin{table}[!ht]
\caption{Results on using different model architectures and augmentation.}\label{ablation_model}
\centering
\begin{tabular}{p{4cm}c}
\toprule
\textbf{Model}	&  \textbf{Performance}\\
\midrule
Plain IndRNN&69.48$\%$\\
Residual IndRNN& 67.98$\%$\\
Dense IndRNN&71.80$\%$\\
Dense-IndRNN-aug& 74.06$\%$\\
\bottomrule
\end{tabular}
\end{table}

Considering the large differences between the training data and validation/test data (from different users), the learned model tends to become overfiting when the learning rate is too small. Therefore, the effects of different learning rates are further studied on the final performance. The results are shown in Fig. \ref{lr}. It can be seen that the network performs similarly in a wide range of learning rates. The learning rate is set to $8\times10^{-5}$ in the experiments.

\subsection{Transfer Learning}

The dense IndRNN model trained above with the feature augmentation and learning rate is used for the transfer learning \cite{transfer} to further improve the performance on the final test dataset as described in Subsection \ref{subsec_trans}. The learning rate in the transfer learning is set to $2\times10^{-5}$ in the training empirically. In this paper, the simple fine-tuning of the model on the transfer learning sets is used. The result is shown in Table \ref{tab_trans}. It can be seen that after transfer learning, the accuracy of validation set increases to $80.72\%$, which means that cross-user transfer learning is useful for testing on the data from different users. It is noticed that the performance of transferB model is better than the transferA, which is due to the uneven distribution of the two transfer learning datasets.

\begin{figure}[!ht]
\centering
\subfigure[The transferA model.]{
\begin{minipage}[!ht]{0.35\linewidth}
\centering
\includegraphics[width=1\linewidth]{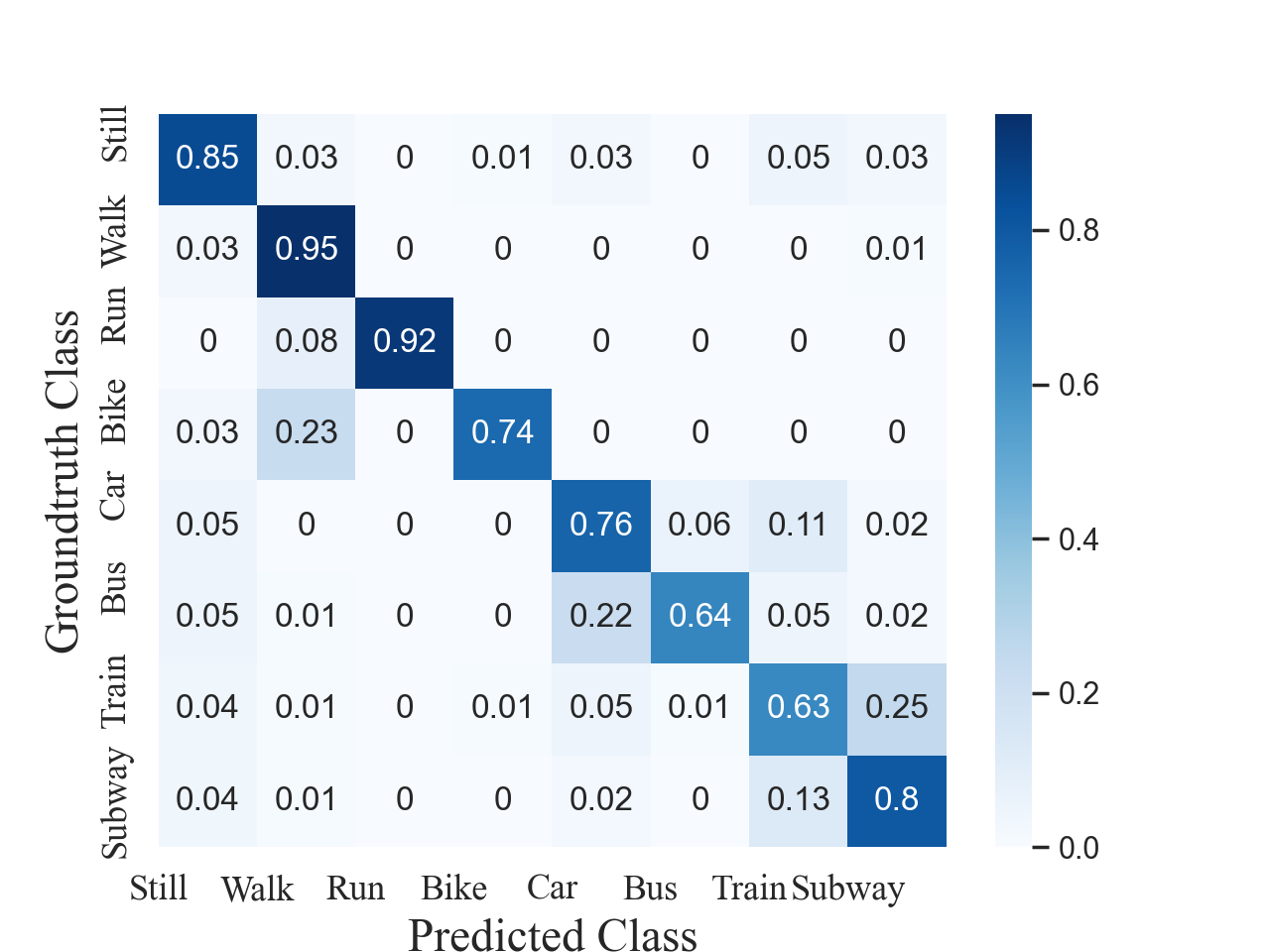}
\vspace{0.005cm}
\end{minipage}%
\hspace{0.10cm}
}%
\subfigure[The transferB model.]{
\begin{minipage}[!ht]{0.33\linewidth}
\centering
\includegraphics[width=1\linewidth]{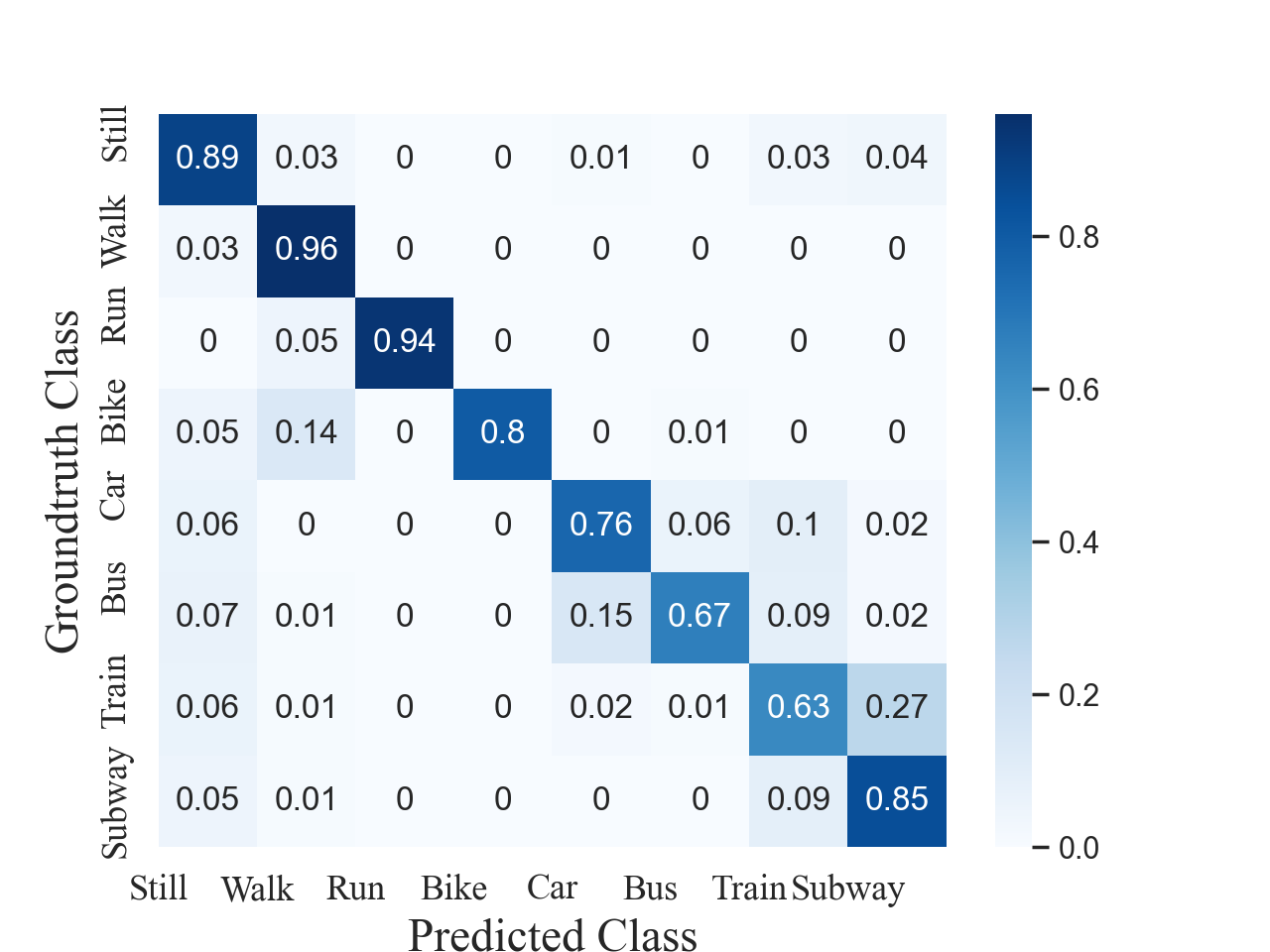}
\vspace{0.005cm}
\end{minipage}%
\hspace{0.05cm}
}%
\subfigure[The final model.]{
\begin{minipage}[!ht]{0.33\linewidth}
\centering
\includegraphics[width=1\linewidth]{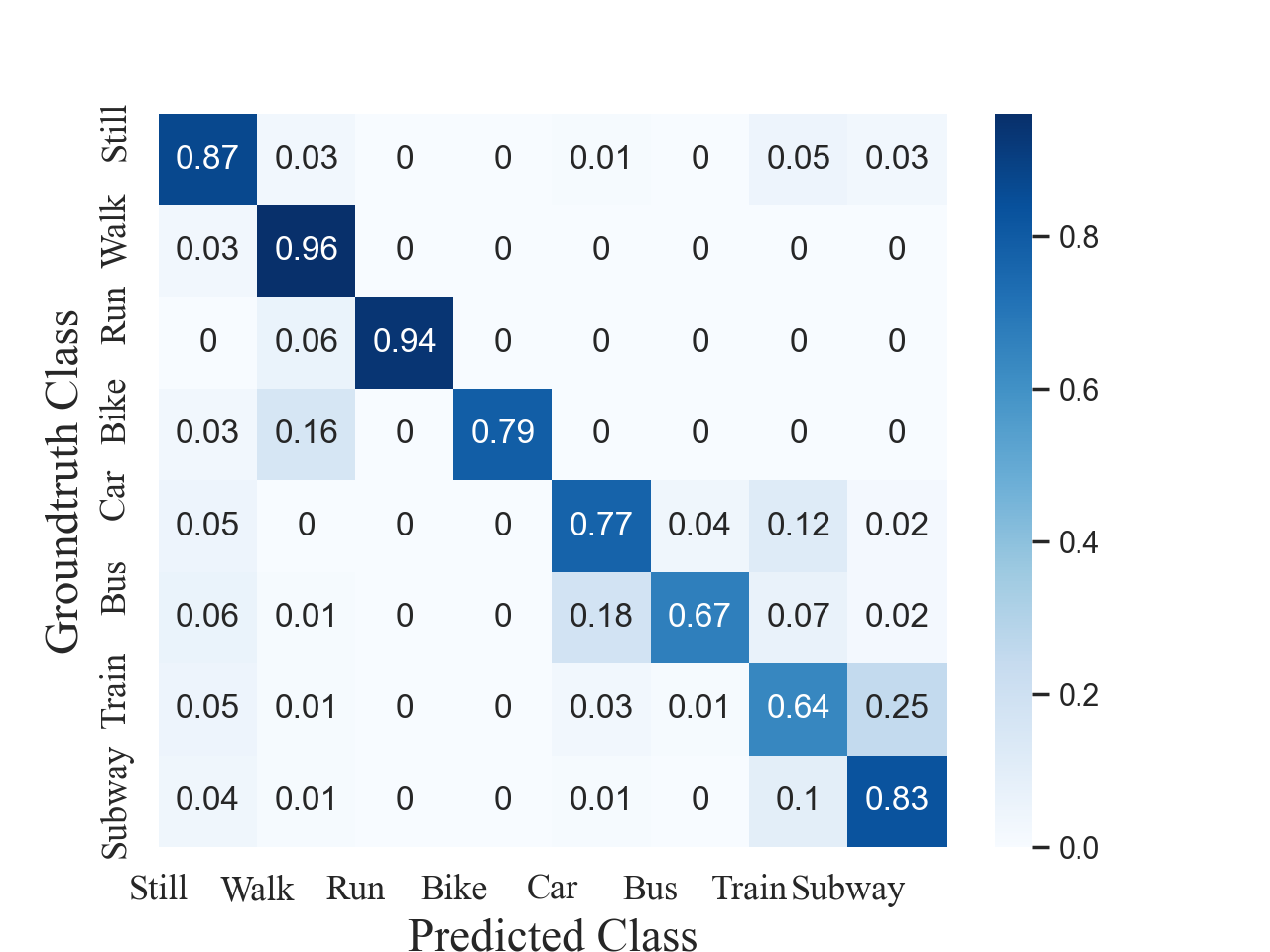}
\vspace{0.005cm}
\end{minipage}%
}%

\centering
\caption{Confusion matrices of the different transfer models.}\label{fig_trans_conf}
\end{figure}
\begin{table}[!ht]
\caption{Results of the different transfer learning models}\label{tab_trans}
\centering
\begin{tabular}{ccc}
\toprule
\textbf{Method}	& \textbf{Performance}&\textbf{Final Performance}\\
\midrule
TransferA		& 78.11$\%$&\multirow{2}*{80.72$\%$}\\
TransferB	&80.97$\%$\\
\bottomrule
\end{tabular}
\end{table}
By comparing the confusion matrices before and after transfer learning shown in Fig. \ref{Confusion of four IndRNN} and Fig. \ref{fig_trans_conf}, it can be further observed that the recognition accuracy increases a lot for most classes (except bike and Bus). For ``Still'', transfer learning further brings an accuracy improvement around 6\%, which eliminates the mistakes of being predicted as ``Bike'', ``Car'' or ``Bus''. For ``Walk'', the accuracy increases around 3\%, mainly reducing the confusion with Train or Subway. Moreover, The accuracy improvement for ``Run'' is significant from 43\% to 94\%. Before transfer learning, over 40$\%$ ``Run'' samples are predicted as ``Bike'', while after that, it has been largely improved. It indicates that the activity ``Run'' is of strong user-dependence. The recognition accuracies of three motor-powered activities, including ``Car'', ``Train'' and ``Subway'', have also been improved while ``Bus'' slightly decreases and misclassified as ``Car''. While the proposed method achieves a relatively high performance on the other locomotions, the accuracies of the four motor-powered activities are still relatively low due to their strong similarities. Methods on distinguishing the small differences among them are highly desired, which will be investigated in the future. 

\subsection{Comparison with State-Of-The-Art Classification Methods}
The proposed method is further compared with the existing methods \cite{No.10,No.11,No.1,No.3,No.4,No.5,No.6,No.7,No.8,No.9,No.14,No.12}. The results are shown in Table \ref{result} including comparisons with the existing machine learning and deep learning methods. It can be seen that the proposed IndRNN long-term temporal recognition greatly improves the performance over other single-model based machine learning and deep learning methods. However, it is slightly worse than the model fusion method of DenseNetX + GRU \cite{No.1} (the first place of the SHL Challenge 2020), which fuses the CNN and RNN models together and also fuses the features of each sensor processed individually. It indicates that the spatial processing and effective combination of all the sensors may be important for the recognition. On the other hand, the proposed IndRNN model can also be equipped with the enhanced spatial processing and combination of sensors to further improve the performance, which will be studied in the future. 

\begin{table}[!ht]
\caption{Results of the proposed method in comparison with the existing methods.}\label{result}
\centering
\begin{tabular}{lc}
\toprule
\textbf{Method}	& \textbf{Performance}\\
\midrule
XGBoost \cite{No.10}& 55.0$\%$\\
Nearest Neighbour smoothing \cite{No.8}& 61.2$\%$\\
Random Forest (without location estimation) \cite{No.7}& 62.6$\%$\\
Random Forest (with location estimation) \cite{No.6} &69.1$\%$\\
XGBoost (semi-supervised) \cite{No.3}&77.9$\%$\\
\midrule
GAN \cite{No.14}& 34.4$\%$\\
Multi-View CNN \cite{No.12} &37.3$\%$\\
Logistic regression \cite{No.9}& 55.7$\%$\\
InceptionTime \cite{No.5}	&69.4$\%$\\
3-Layer CNN \cite{No.4}		&76.4$\%$ \\
\midrule
CNN + LSTM \cite{No.11}&52.8$\%$\\
DenseNetX + GRU (Model Fusion based) \cite{No.1}& 88.5$\%$\\
\midrule
{\bf Dense-IndRNN-aug}	& 80.7$\%$\\
\bottomrule
\end{tabular}
\end{table}

\section{Conclusion}
In this paper,  we have presented a framework of combining short-term spatial/frequency feature extraction and long-term IndRNN model for smartphone sensors based activity recognition. The short-term spatial and frequency domain features are first extracted with the Fourier transform to deal with the periodic nature of the sensor data. Together with the conventional statistical features, the FFT amplitude spectrums and the statistical features of the FFT spectrums are extracted to characterize the data of a short-term window. Then a dense IndRNN model is further developed to learn the long-term temporal features on top of the short-term spatial and frequency domain features. Finally, transfer learning is adopted in the experiments to realize the user-independence, which further improves the performance on the test set. Experiments show that our model achieved an accuracy of 80.72 $\%$ on the SHL dataset, which is better than the existing single-model based methods.
\vspace{6pt} 

\authorcontributions{Formal analysis, Beidi Zhao, Shuai Li, Yanbo Gao and Chuankun Li; Methodology, Beidi Zhao, Shuai Li and Wanqing Li; Software, Beidi Zhao, Yanbo Gao and Chuankun Li; Supervision, Shuai Li; Writing – original draft, Beidi Zhao; Writing – review \& editing, Shuai Li, Wanqing Li. All authors have read and agreed to the published version of the manuscript.}

\funding{This work was supported by the National Key R\&D Program of China (2018YFE0203900) and the National Natural Science Foundation of China (No.61901083 and No. 62001092).}


\conflictsofinterest{The authors declare no conflict of interest.} 

%


\reftitle{References}



\begin{thebibliography}{999}

\bibitem[Author1(year)]{video surveillance1}
Vishwakarma, S.; Agrawal, A. A survey on activity recognition and behavior understanding in video surveillance. {\em Vis Comput.} {\bf 2013}, 983–1009.

\bibitem[Author1(year)]{video surveillance2}
Lin, W.; Sun, M.; Poovandran, R.; Zhang, Z. Human activity recognition for video surveillance. {\em IEEE International Symposium on Circuits and Systems} {\bf 2008}, 2737-2740.

\bibitem[Author1(year)]{video surveillance3}
Niu, W.; Long, J.; Han, D.; Wang, Y. Human activity detection and recognition for video surveillance. {\em IEEE International Conference on Multimedia and Expo (ICME)} {\bf 2004}, {\em 1}, 719-722. 

\bibitem[Author1(year)]{human-computer interaction}
Ruffieux, S.; Lalanne, D.; Mugellini, E. ChAirGest: a challenge for multimodal mid-air gesture recognition for close HCI. {\em Proceedings of the 15th ACM on International conference on multimodal interaction (ICMI '13)} {\bf 2013}, 483–488.

\bibitem[Author1(year)]{gaming}
Biancat, J.; Brighenti, C.; Brighenti, A. Review of transportation mode detection techniques. {\em Proceedings of the IEEE Computer Society Conference on Computer Vision and Pattern Recognition (CVPR) Workshops} {\bf 2014}, 7-12.

 \bibitem[Author1(year)]{general0}
Hassana, M.M.; Uddin, M.Z.; Mohamed, A.; Almogren, A. A robust human activity recognition system using smartphone sensors and deep learning. {\em Future Generation Computer Systems} {\bf 2018}, {\em 81}, 307-313.

\bibitem[Author1(year)]{general}
Biancat, J.; Brighenti, C.; Brighenti, A. Review of transportation mode detection techniques. {\em EAI Endorsed Transactions on Ambient Systems} {\bf 2014}, {\em 1}, 1-10.

\bibitem[Author1(year)]{general1}
Ravì, D.; Wong, C.; Lo, B.; Yang, G. A Deep Learning Approach to on-Node Sensor Data Analytics for Mobile or Wearable Devices. {\em IEEE Journal of Biomedical and Health Informatics} {\bf 2017}, {\em 21}, 56-64.

\bibitem[Author1(year)]{Smartphone-Based}
Zhou, B.; Yang, J.; Li, Q. Smartphone-Based activity recognition for indoor localization using a convolutional neural network. {\em Sensors} {\bf 2019}, {\em 19}, 621.

\bibitem[Author1(year)]{Real-time}
Shotton, J.; Fitzgibbon A.; Cook, M.; Sharp, T.; Finocchio, M.; Moore, R.; Kipman, A.; Blake, A. Real-time human pose recognition in parts from single depth images. {\em Proceedings of the IEEE Conference on Computer Vision and Pattern Recognition (CVPR)} {\bf 2011}, 1297--1304.

\bibitem[Author1(year)]{convention_video}
Peng, B.; Lei, J.; Fu, H.; Shao, L.; Huang, Q. A Recursive Constrained Framework for Unsupervised Video Action Clustering. {\em IEEE Transactions on Industrial Informatics} {\bf 2020}, 555-565.

\bibitem[Author1(year)]{frontier}
Vito, J.; Rešçiç, N.; Mlakar, M.; Drobnič, V.; Gams, M.; Slapničar, G.; Gjoreski, M.; Bizjak, J;. Marinko, M.; Luštrek, M. A New Frontier for Activity Recognition: The Sussex-Huawei Locomotion Challenge. {\em Proceedings of the 2018 ACM International Joint Conference and 2018 International Symposium on Pervasive and Ubiquitous Computing and Wearable Computers (UbiComp '18)} {\bf 2018}, 1511-1520.

\bibitem[Author1(year)]{summary_2020}
Wang, L.; Gjoreski, H.; Ciliberto, M.; Lago, P.; Murao, K.; Okita, T.; Roggen, D. Summary of the Sussex-Huawei Locomotion-Transportation Recognition Challenge 2020. {\em Proceedings of the 2020 ACM International Joint Conference and 2020 International Symposium on Pervasive and Ubiquitous Computing and Wearable Computers} {\bf 2020}, 351-358.

\bibitem[Author1(year)]{dataset1}
Wang, L.; Gjoreski, H.; Ciliberto, M.; Mekki, S.; Valentin, S.; Roggen, D. Enabling reproducible research in sensor-based transportation mode recognition with the Sussex-Huawei dataset. {\em IEEE Access} {\bf 2019}, {\em 7}, 10870-10891.

\bibitem[Author1(year)]{EmbraceNet}
Choi, J.; Lee, J. EmbraceNet for activity: a deep multimodal fusion architecture for activity recognition. {\em Proceedings of the 2019 ACM International Joint Conference on Pervasive and Ubiquitous Computing and Proceedings of the 2019 ACM International Symposium on Wearable Computers (UbiComp/ISWC '19 Adjunct)} {\bf 2019}, 693–698.

\bibitem[Author1(year)]{1D DenseNet}
Zhu, Y.; Zhao, F.; Chen, R. Applying 1D sensor DenseNet to Sussex-Huawei locomotion-transportation recognition challenge. {\em Adjunct Proceedings of the 2019 ACM International Joint Conference on Pervasive and Ubiquitous Computing and Proceedings of the 2019 ACM International Symposium on Wearable Computers (UbiComp/ISWC '19 Adjunct)} {\bf 2019}, 873-877.

\bibitem[Author1(year)]{DCLSTMMWAR}
Ordóñez, F. J.; Roggen, D. Deep Convolutional and LSTM Recurrent Neural Networks for Multimodal Wearable Activity Recognition. {\em Sensors} {\bf 2016}, {\em 16}, 115.

\bibitem[Author1(year)]{No.2}
Zhao, B.; Li, S.; Gao, Y. IndRNN based long-term temporal recognition in the spatial and frequency domain.  {\em Proceedings of the 2020 ACM International Joint Conference on Pervasive and Ubiquitous Computing and Proceedings of the 2020 ACM International Symposium on Wearable Computers (UbiComp-ISWC '20)} {\bf 2020}, 368-372.

\bibitem[Author1(year)]{nurse}
Inoue, S.; Ueda, N.;  Nohara, Y.; Nakashima, N. Mobile activity recognition for a whole day: recognizing real nursing activities with big dataset. {\em Proceedings of the 2015 ACM International Joint Conference on Pervasive and Ubiquitous Computing (UbiComp '15)} {\bf 2015}, 1269–1280.

\bibitem[Author1(year)]{RTSACUISR}
Zhuo, S.; Sherlock, L.; Dobbie, G.; Koh, Y.S.; Russello, G.; Lottridge, D. Real-time smartphone activity classification using inertial sensors—recognition of scrolling, typing, and watching videos while sitting or walking. {\em Sensors} {\bf 2020}, {\em 20}, 655.

\bibitem[Author1(year)]{ACURDWS}
Parkka, J.; Ermes, M.; Korpipaa, P.; Mantyjarvi, J.; Peltola, J.; Korhonen, I. Activity classification using realistic data from wearable sensors. {\em IEEE Transactions on Information Technology in Biomedicine}, {\bf 2006}, {\em 10}, 119-128.

\bibitem[Author1(year)]{UDOIHARMP}
Ustev, Y.E.; Incel, O.D.; Ersoy, C. User, device and orientation independent human activity recognition on mobile phones: Challenges and a proposal. {\em Proceedings of the 2013 ACM conference on Pervasive and ubiquitous computing adjunct publication, Zurich, Switzerland} {\bf 2013}, 1427–1436.

\bibitem[Author1(year)]{HMC}
Kim, Y.; Kang, B.; Kim, D. Hidden Markov Model Ensemble for Activity Recognition Using Tri-Axis Accelerometer. {\em IEEE International Conference on Systems, Man, and Cybernetics} {\bf 2015}, 3036-3041.

\bibitem[Author1(year)]{SVM}
Cortes, C. and Vapnik, V. Support vector networks. {\em Machine Learning} {\bf 1995}, {\em 20}, 273-297.

\bibitem[Author1(year)]{SVM1}
Fleury, A.; Vacher, M.;  Noury, N. SVM-Based Multimodal Classification of Activities of Daily Living in Health Smart Homes: Sensors, Algorithms, and First Experimental Results. {\em IEEE Transactions on Information Technology in Biomedicine} {\bf 2010}, {\em 14}, 274-283.

\bibitem[Author1(year)]{Random Forest+HMM}
Janko, V.; Gjoreski, M.; De Masi, C.M.; Reščič, N.; Luštrek, M.; Gams, M. Cross-location transfer learning for the sussex-huawei locomotion recognition challenge. {\em Proceedings of the 2019 ACM International Joint Conference on Pervasive and Ubiquitous Computing and Proceedings of the 2019 ACM International Symposium on Wearable Computers (UbiComp/ISWC '19 Adjunct)} {\bf 2019}, 730-735.

\bibitem[Author1(year)]{CNNHARUMS}
Zeng, M.; Nguyen, L.T.; Yu, B.; Mengshoel, O. J.; Zhu, J.; Wu, P.; Zhang, J. Convolutional neural networks for human activity recognition using mobile sen- sors.  {\em Proceedings of international conference on mobile computing, applications and services (MobiCASE)} {\bf 2013}, 197–205.

\bibitem[Author1(year)]{TSCUMDCNN}
Zheng, Y.; Liu, Q.; Chen, E.; Ge; Y.; Zhao, J.L. Time series classification using multi-channels deep convolutional neural networks. {\em Web-age informa- tion management} {\bf 2014}, {\em 8485}, 298-310.

\bibitem[Author1(year)]{HARSSUDLNN}
Ronao, C.A.; Cho, S.B. Human activity recognition with smartphone sensors using deep learning neural networks. {\em Expert Systems with Applications}, {\bf 2016}, {\em 59}, 235-244.

\bibitem[Author1(year)]{deepLSTM}
Pradhan, S.; Longpre, S. Exploring the depths of recurrent neural networks with stochastic residual learning. Report.

\bibitem[Author1(year)]{RNN1}
Xi, R.; Li, M.; Hou, M.; Fu, M.; Qu, M.; Liu, D.; Haruna, C.R. Deep Dilation on Multimodality Time Series for Human Activity Recognition. {\em IEEE Accesss} {\bf 2018}, {\em 6}, 53381-53396.

\bibitem[Author1(year)]{IndRNN2018}
Li, S.; Li, W.; Cook, C.; Zhu, C.; Gao, Y. Independently Recurrent Neural Network (IndRNN): Building A Longer and Deeper RNN. {\em Proceedings of the IEEE Conference on Computer Vision and Pattern Recognition (CVPR)} {\bf 2018}, 5457–5466.

\bibitem[Author1(year)]{IndRNN2019}
Li, S.; Li, W.; Cook, C.; Gao, Y. Deep Independently Recurrent Neural Network (IndRNN). {\bf 2019}, arXiv.cs.CV 1910.06251.

\bibitem[Author1(year)]{IndRNN_for_SHL_2018}
Li,S.; Li, C.; Li, W.; Hou, Y.; Cook, C. Smartphone-sensors Based Activity Recognition Using IndRNN. {\em Proceedings of the 2018 ACM International Joint Conference and 2018 International Symposium on Pervasive and Ubiquitous Computing and Wearable Computers (UbiComp '18)} {\bf 2018}, 1541-1547.

\bibitem[Author1(year)]{IndRNN_for_SHL_2019}
Zheng, L.; Li, S.; Zhu, C.; Gao, Y. Application of IndRNN for human activity recognition: the Sussex-Huawei locomotion-transportation challenge. {\em Proceedings of the 2019 ACM International Joint Conference on Pervasive and Ubiquitous Computing and Proceedings of the 2019 ACM International Symposium on Wearable Computers (UbiComp/ISWC '19 Adjunct)}. 869–872.

\bibitem[Author1(year)]{dataset2}
Gjoreski, H.; Ciliberto, M.; Wang, L.;  Morales, F.J.O.; Mekki, S.; Valentin, S.; Roggen, D. The University of Sussex-Huawei locomotion and transportation dataset for multimodal analytics with mobile devices. {\em IEEE Access} {\bf 2018}, 42592-42604.

\bibitem[Author1(year)]{Adam}
Diederik, K; Jimmy, B. Adam: A Method for Stochastic Optimization. {\em International Conference on Learning Representations.} {\bf 2014}.

\bibitem[Author1(year)]{transfer}
Zhang, J.; Li, W.; Ogunbona, P. Joint Geometrical and Statistical Alignment for Visual Domain Adaptation. {\em Proceedings of the IEEE Conference on Computer Vision and Pattern Recognition (CVPR)}  {\bf 2017}, 1859-1867.

\bibitem[Author1(year)]{No.1}
Zhu, Y.; Luo, H.; Chen, R.; Zhao, F.; Su, L. DenseNetX and GRU for the sussex-huawei locomotion-transportation recognition challenge. {\em Proceedings of the 2020 ACM International Joint Conference on Pervasive and Ubiquitous Computing and Proceedings of the 2020 ACM International Symposium on Wearable Computers (UbiComp-ISWC '20)} {\bf 2020}, 373–377.

\bibitem[Author1(year)]{No.3}
Kalabakov, S.; Stankoski, S.; Reščič, N.; Kiprijanovska, I.; Andova, A.; Picard, C.; Janko, V.; Gjoreski, M.; Luštrek, M. Tackling the SHL challenge 2020 with person-specific classifiers and semi-supervised learning. {\em Proceedings of the 2020 ACM International Joint Conference on Pervasive and Ubiquitous Computing and Proceedings of the 2020 ACM International Symposium on Wearable Computers (UbiComp-ISWC '20)} {\bf 2020}, 323–328.

\bibitem[Author1(year)]{No.4}
Yaguchi, K.; Ikarigawa, K.; Kawasaki, R.; Miyazaki, W.; Morikawa, Y.; Ito, C.; Shuzo, M.; Maeda, E. Human activity recognition using multi-input CNN model with FFT spectrograms.{\em Proceedings of the 2020 ACM International Joint Conference on Pervasive and Ubiquitous Computing and Proceedings of the 2020 ACM International Symposium on Wearable Computers (UbiComp-ISWC '20)} {\bf 2020}, 364–367.

\bibitem[Author1(year)]{No.5}
Naseeb, C.; Saeedi, B.A. Activity recognition for locomotion and transportation dataset using deep learning. {\em Proceedings of the 2020 ACM International Joint Conference on Pervasive and Ubiquitous Computing and Proceedings of the 2020 ACM International Symposium on Wearable Computers (UbiComp-ISWC '20)} {\bf 2020}, 329-334.

\bibitem[Author1(year)]{No.6}
Siraj, M.S.; Faisal, M.A.A.; Shahid, O.; Abir, F.F.; Hossain, T.; Inoue, S.; Ahad, M.A.R. UPIC: user and position independent classical approach for locomotion and transportation modes recognition. {\em Proceedings of the 2020 ACM International Joint Conference on Pervasive and Ubiquitous Computing and Proceedings of the 2020 ACM International Symposium on Wearable Computers (UbiComp-ISWC '20)} {\bf 2020}, 340-345.

\bibitem[Author1(year)]{No.7}
Brajesh, S.; Ray, I. Ensemble approach for sensor-based human activity recognition.{\em Proceedings of the 2020 ACM International Joint Conference on Pervasive and Ubiquitous Computing and Proceedings of the 2020 ACM International Symposium on Wearable Computers (UbiComp-ISWC '20)} {\bf 2020}, 296-300.

\bibitem[Author1(year)]{No.8}
Widhalm, P.; Merz, P.; Coconu, L.; Brändle, N. Tackling the SHL recognition challenge with phone position detection and nearest neighbour smoothing.{\em Proceedings of the 2020 ACM International Joint Conference on Pervasive and Ubiquitous Computing and Proceedings of the 2020 ACM International Symposium on Wearable Computers (UbiComp-ISWC '20)} {\bf 2020}, 359-363.

\bibitem[Author1(year)]{No.9}
Sekiguchi, R.; Abe, K.; Yokoyama, T.; Kumano, M.; Kawakatsu, M. Ensemble learning for human activity recognition. {\em Proceedings of the 2020 ACM International Joint Conference on Pervasive and Ubiquitous Computing and Proceedings of the 2020 ACM International Symposium on Wearable Computers (UbiComp-ISWC '20)} {\bf 2020}, 335-339.

\bibitem[Author1(year)]{No.10}
Tseng, Y.; Lin, H.; Lin, Y.; Chen, J. Hierarchical classification using ML/DL for sussex-huawei locomotion-transportation (SHL) recognition challenge. {\em Proceedings of the 2020 ACM International Joint Conference on Pervasive and Ubiquitous Computing and Proceedings of the 2020 ACM International Symposium on Wearable Computers (UbiComp-ISWC '20)} {\bf 2020}, 346-350.

\bibitem[Author1(year)]{No.11}
Friedrich, B.; Lübbe, C.; Hein, A. Combining LSTM and CNN for mode of transportation classification from smartphone sensors. {\em Proceedings of the 2020 ACM International Joint Conference on Pervasive and Ubiquitous Computing and Proceedings of the 2020 ACM International Symposium on Wearable Computers (UbiComp-ISWC '20)} {\bf 2020}, 305–310.

\bibitem[Author1(year)]{No.12}
Hamidi, M.; Osmani, A.; Alizadeh, P. A multi-view architecture for the SHL challenge. {\em Proceedings of the 2020 ACM International Joint Conference on Pervasive and Ubiquitous Computing and Proceedings of the 2020 ACM International Symposium on Wearable Computers (UbiComp-ISWC '20)} {\bf 2020}, 317–322.

\bibitem[Author1(year)]{No.14}
Gunthermann, L.; Simpson, I.; Roggen, D. Smartphone location identification and transport mode recognition using an ensemble of generative adversarial networks. {\em Proceedings of the 2020 ACM International Joint Conference on Pervasive and Ubiquitous Computing and Proceedings of the 2020 ACM International Symposium on Wearable Computers (UbiComp-ISWC '20)} {\bf 2020}, 311–316.



\end{thebibliography}



\end{document}